\documentclass[10pt,twocolumn,letterpaper]{article}

\usepackage{cvpr}              

\usepackage{graphicx}
\usepackage{amsmath}
\usepackage{amssymb}
\usepackage{booktabs}
\usepackage{makecell}
\usepackage{bm}
\usepackage{bbm}
\usepackage{mathtools}
\usepackage{enumitem}

\DeclareMathOperator{\onehot}{one\_hot}
\DeclareMathOperator{\mean}{mean}
\DeclareMathOperator{\argmax}{argmax}

\usepackage[pagebackref,breaklinks,colorlinks]{hyperref}

\usepackage[capitalize]{cleveref}
\crefname{section}{Sec.}{Secs.}
\Crefname{section}{Section}{Sections}
\Crefname{table}{Table}{Tables}
\crefname{table}{Tab.}{Tabs.}

\def\cvprPaperID{2976} 
\def\confName{CVPR}
\def\confYear{2022}

\begin{document}

\title{Robust Mutual Learning for Semi-supervised Semantic Segmentation}


\author{%
  Pan Zhang$^1$\thanks{Work was done during the first author's internship at Microsoft Research Asia.}, Bo Zhang$^2$, Ting Zhang$^2$, Dong Chen$^2$, Fang Wen$^2$ \\
  $^1$University of Science and Technology of China, $^2$Microsoft Research Asia\\
  {\tt\small zhangpan@mail.ustc.edu.cn} \quad
  {\tt\small \{zhanbo, tinzhan, doch, fangwen\}@microsoft.com} \\
}
\maketitle

\begin{abstract}
   Recent semi-supervised learning (SSL) methods are commonly based on pseudo labeling.  Since the SSL performance is greatly influenced by the quality of pseudo labels, mutual learning has been proposed to effectively suppress the noises in the pseudo supervision.
  In this work, we propose robust mutual learning that improves the prior approach in two aspects. First, the vanilla mutual learners suffer from the coupling issue that models may converge to learn homogeneous knowledge. We resolve this issue by introducing mean teachers to generate mutual supervisions so that there is no direct interaction between the two students. We also show that strong data augmentations, model noises and heterogeneous network architectures are essential to alleviate the model coupling. Second, we notice that mutual learning fails to leverage the network’s own ability for pseudo label refinement.
  Therefore, we introduce self-rectification that leverages the internal knowledge and explicitly rectifies the pseudo labels before the mutual teaching. Such self-rectification and mutual teaching collaboratively improve the pseudo label accuracy throughout the learning. The proposed robust mutual learning demonstrates state-of-the-art performance on semantic segmentation in low-data regime.
\end{abstract}

\section{Introduction}
\label{sec:intro}

\begin{figure*}[t]
	\center
	\small
	\includegraphics[width=1.90\columnwidth]{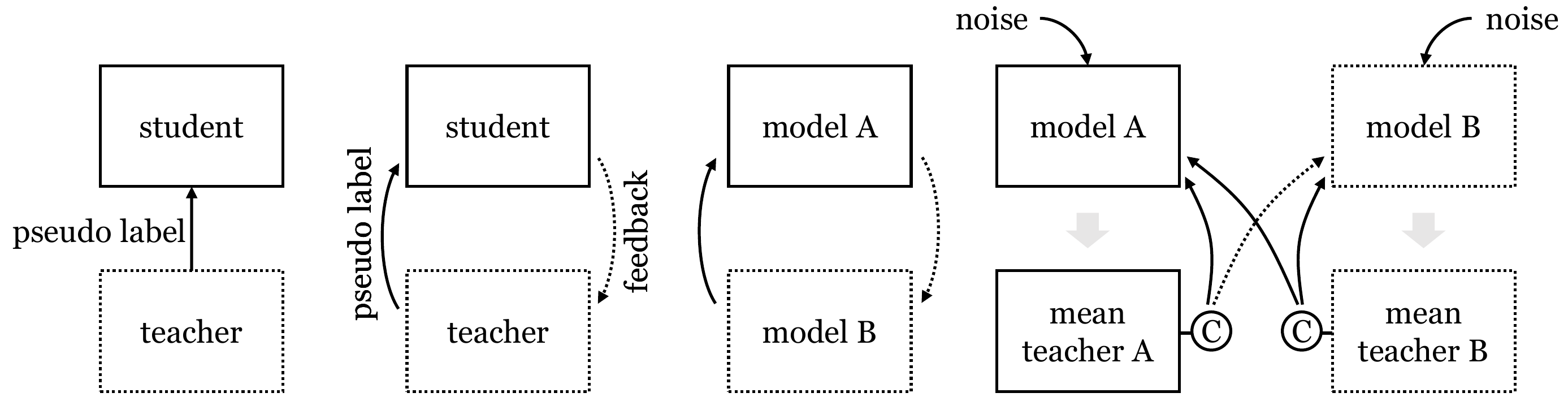}
	\put(-405,-8){\footnotesize (a)}
    \put(-316,-8){\footnotesize (b)}
    \put(-222,-8){\footnotesize (c)}
    \put(-95,-8){\footnotesize (d)}
	\caption{Comparison of (a) pseudo labeling~\cite{lee2013pseudo}, (b) meta pseudo labeling~\cite{pham2020meta}, (c) mutual learning~\cite{ke2020guided,fenga2020dmt} and  (d) the proposed robust mutual learning. The symbol \textcopyright \ denotes the pseudo label self-rectification that explicitly online refines the pseudo label supervisions.}
	\label{fig:diagram}
	\vspace{-1.0em}
\end{figure*}
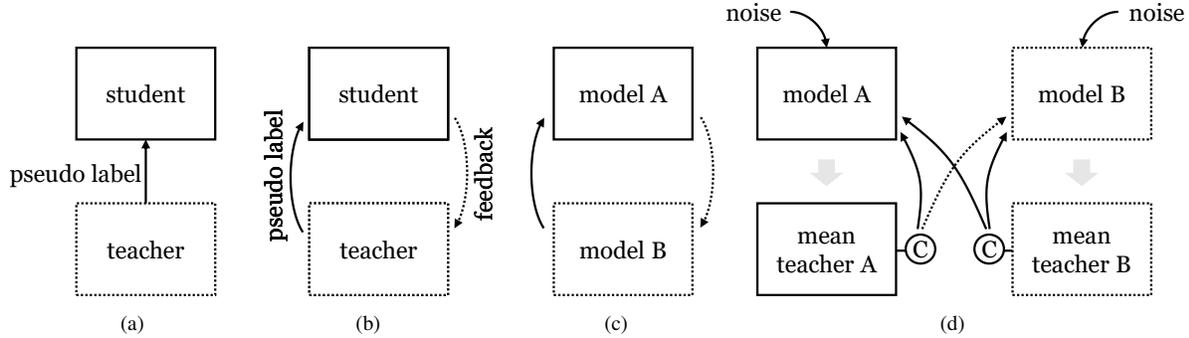

Semi-supervised learning~\cite{miyato2018virtual,sajjadi2016regularization,laine2016temporal,tarvainen2017mean,xie2019unsupervised,verma2019interpolation} improves the data efficiency for low-data regime relying on limited amounts of labeled images and extra massive amounts of unlabeled images. 
Recent state-of-the-art methods, \eg, MeanTeacher~\cite{tarvainen2017mean}, FixMatch~\cite{sohn2020fixmatch} and Noisy student~\cite{xie2020self}, share a similar philosophy and enjoy the merit of both \emph{pseudo labeling}~\cite{lee2013pseudo,xie2020self,pham2020meta} and \emph{consistency regularization}~\cite{sajjadi2016regularization,laine2016temporal,tarvainen2017mean,miyato2018virtual,yun2019cutmix}: the teacher model generates pseudo labels for weakly-augmented unlabeled data whereas the student model trains on the strong-augmented counterparts, as shown in Figure~\ref{fig:diagram}a. Such strategy can also be applied to semantic segmentation, and the resulting approaches~\cite{zou2020pseudoseg,french2020semi,olsson2021classmix,yuan2021simple,ouali2020semi,ke2020three,ke2020guided,fenga2020dmt} have demonstrated outstanding performance.

However, pseudo labeling methods are plagued with confirmation bias~\cite{arazo2020pseudo}, \ie, the student model is prone to overfit the erroneous pseudo labels. A few recent works set out to address this issue, either by estimating the pseudo label uncertainty~\cite{zheng2021rectifying,rizve2021defense} or directly rectifying the pseudo labels~\cite{mendel2020semi,zhang2021prototypical}. Recent work~\cite{pham2020meta} offers breathtaking performance on ImageNet benchmark and the key is to inform the teacher of the feedback from the student model so that better pseudo labels can be generated (shown in Figure~\ref{fig:diagram}b). Mutual learning~\cite{ke2020guided,fenga2020dmt}, on the other hand, trains two models in parallel that play the dual role of teacher and student (shown in Figure~\ref{fig:diagram}c). The networks suppress the noisy knowledge of the counterpart and collaboratively improve in the course of learning. Nonetheless, both networks may end up with learning homogeneous knowledge and suffer from coupled noises that hinder the further improvement of the two learners. Besides, the pseudo label noises are suppressed by purely relying on the knowledge from the peer network, while totally overlooking the network's own ability to improve the pseudo labels. 

In this work, we propose \emph{robust mutual learning} that ameliorates the \emph{coupling issue} of mutual learning and introduce \emph{self-rectification} for the pseudo label refinement.
First, we observe that it is crucial to ensure heterogeneous knowledge during training so that the model coupling can be reduced.
We comprehensively examine the factors to alleviate the issue:
1) we compute the pseudo labels from the mean teacher~\cite{tarvainen2017mean} which not only yields more reliable pseudo labels but also reduces the coupling issue since the learners no longer interact directly;
2) we further explore factors that reduce the coupling of the mutual learners and find that strong data augmentations and model noises (\eg, dropout~\cite{srivastava2014dropout} and stochastic depth~\cite{huang2016deep}) are essential to let models learn incoherent knowledge;
3) we also investigate heterogeneous network architectures (\eg, convolutional neural network versus transformer~\cite{dosovitskiy2020image,zheng2020rethinking}) and demonstrate its effectiveness.

Meanwhile, besides the refinement from the peer network, we propose to explicitly denoise the pseudo labels prior to the mutual learning so that only robust knowledge is propagated throughout the framework. Specifically, we draw inspiration from ~\cite{zhang2021prototypical} and propose to refine pseudo labels according to the 
class-wise prediction confidence estimated based on the relative feature distances to the class centroids. Figure~\ref{fig:diagram} shows the comparison of our framework and prior works. Our framework has two distinct schemes for label refinement: self-rectification leverages the model's own knowledge to refine the pseudo labels and mutual teaching filters the noisy knowledge with the help of the heterogeneous peer network. Both schemes collaboratively improve the pseudo labels during training, leading to the SSL models with narrowed gap to supervised learning. 

Extensive experiments show that the proposed robust mutual learning outperforms prior competitive approaches. On the Cityscapes dataset, our method is best performed under all the SSL settings. In particular, our method shows much superiority on extreme low data regime. When using only $1/30$ labeled images, our method outperforms the state-of-the-art method~\cite{fenga2020dmt} by $3.45\%$ mIoU. Our method also demonstrates performance advantage on the PASCAL VOC 2012 dataset. Notably, our method could offer performance on-par with supervised learning merely using $1/8$ labeled data.

Our main contribution can be summarized as follows. 1) We notice the coupling issue of existing mutual learning works and exhaustively examine factors that can be used to ameliorate that issue, leading to improved performance thanks to the induced heterogeneous knowledge learning. 2) We propose to leverage the model's internal knowledge besides the external knowledge from peer networks and the pseudo labels are further rectified according to the feature distances to class centroids,
boosting the performance one step further to a record high. 3) The proposed robust mutual learning could better online refine the pseudo labels and achieves state-of-the-art performance on several commonly used semantic segmentation datasets. 

\section{Related Works}

The key of semi-supervised segmentation lies in how to leverage the large unlabeled data set to derive additional training signals.
Along this direction, prior works can be roughly categorized into five classes:
generative learning, contrastive learning, entropy minimization, consistency regularization and pseudo labeling.
We briefly review the last two categories which are mostly related and widely incorporated in current state-of-the-art methods.

\noindent {\bf Consistency regularization.}
Consistency regularization enforces the model to make consistent predictions with respect to various perturbations.
The effectiveness of consistency regularization is based on the smoothness assumption or the cluster assumption that data points close to each other are likely to be from the same class,
which is often held in classification tasks and has brought about a lot of research efforts~\cite{xie2019unsupervised,sohn2020fixmatch,berthelot2019remixmatch,sajjadi2016regularization,tarvainen1780weight,berthelot2019mixmatch,kim2021selfmatch,li2018semi,perone2018deep}.
As for semantic segmentation, 
it is observed in~\cite{french2020semi,ouali2020semi} that the cluster assumption is violated in the semantic segmentation task.
Therefore, Ouali et al.~\cite{ouali2020semi} propose to perturb the encoder's outputs where the cluster assumption is maintained and leverage multiple auxiliary decoders to give consistent predictions.
French et al.~\cite{french2020semi} find that mask-based augmentation strategies are effective
and introduce an adapted version of a popular technique CutMix~\cite{french2020semi}.
The idea of CutMix is to mix samples by replacing the image region with a patch from another image, which can be regarded as an extension of Cutout~\cite{devries2017improved} and Mixup~\cite{zhang2017mixup}, 
and is further extended in recent works~\cite{olsson2021classmix, chen2021mask,french2019consistency}.
Our approach also adopts the idea from CutMix~\cite{french2020semi} to enforce consistency between the mixed outputs and the prediction over the mixed inputs.

\noindent {\bf Pseudo labeling.}
Pseudo labeling or self-training is a typical technique in leveraging unlabeled data by alternating the pseudo label prediction and feature learning, which encourages the model to make confident predictions for unlabeled data. This technique has been successfully and widely applied to many tasks such as natural language processing~\cite{yarowsky1995unsupervised,he2019revisiting, kahn2020self, park2020improved}, image classification~\cite{lee2013pseudo,zhai2019s4l,zoph2020rethinking,xie2020self,yalniz2019billion} as well as semantic segmentation~\cite{feng2020semi,chen2020leveraging,zoph2020rethinking}.
The key of pseudo labeling relies on the quality of the pseudo labels.
Most models~\cite{lee2013pseudo,tarvainen2017mean,sohn2020fixmatch,xie2020self} refine the pseudo label from external guidance, \eg, a teacher model.
However, the teacher model is usually fixed, making the student inherit some inaccurate predictions from the teacher.
Recent efforts turn to update the teacher along with the student in order to generate better pseudo labels, \eg, co-teaching~\cite{yu2019does,han2018co}, dual student~\cite{ke2019dual}, meta pseudo label~\cite{pham2020meta}, mutual training~\cite{ke2020guided,fenga2020dmt} and so on.
Such mutual learning strategy however may end up learning homogeneous knowledge and hence fail to provide complementary information for each other. 

In this paper, we point out that learning heterogeneous knowledge in mutual learning is of great importance but is rarely noticed.
A close related work~\cite{ge2020mutual} also adopts mean teaching for unsupervised domain adaptation in person re-identification. In contrast, our work differs in two aspects: 1) we observe the importance of learning heterogeneous knowledge in semi-supervised learning and mean teaching is one of the proposed potential solutions;
2) besides mutual pseudo label refinement, we also explore internal guidance to improve the quality of the pseudo label.

\section{Preliminary}
Semi-supervised learning (SSL) for semantic segmentation learns from the pixel-wise labeled dataset $\mathcal{D}_l=\{\bm{x}_l, \bm{y}_l\}_{l=1}^{n_l}$ in conjunction with unlabeled data $\mathcal{D}_u=\{\bm{x}_u\}_{u=1}^{n_u}$, where $\bm{x}\in \mathbb{R}^{H\times W \times 3}$ denotes the training images with resolution of $H\times W$ and $\bm{y}_l\in \mathbb{R}^{H\times W \times K}$ is the ground truth corresponding to $\bm{x}_l$ with pixels labeled by $K$ classes. The resulting segmentation network $h_{\bm{\theta}}$ parameterized by ${\bm{\theta}}$ can be regarded as a composite of a feature extractor $f$ and a linear classifier $g$, \ie, $h= g \odot f$.

Recent state-of-the-art approaches follow a similar paradigm: 
the network is first trained on the labeled images with weak augmentations $\gamma$ by minimizing the standard cross-entropy, \ie,
\begin{equation}
  \mathcal{L}_s({\bm{\theta}}) = H((\bm{y}_l, h_{\bm{\theta}}(\gamma(\bm{x}_l))).
  \label{eq:supervised}
\end{equation}
Here $\gamma$ is photometric transformation, and the geometric transformation can also be both additionally adopted to the labels and images. Then, for unlabeled images the ``hard'' pseudo labels\footnote{Recent works find hard pseudo labels can lead to stronger performance. Our experiment in the supplementary material also proves this. } $\hat{\mathbf{y}}$ can be generated, which is a 
one-hot vector converted from the soft prediction, \ie, $\hat{\mathbf{y}} = \onehot( h_\theta(\gamma(\mathbf{x}_u)))$. The prediction for the strongly augmented images should match the pseudo labels. Also, the unreliable pseudo labels are discounted when the confidence falls below a predefined threshold $\tau$, so the unsupervised objective for unlabeled images can be formulated as:
\begin{equation}
  \mathcal{L}_u({\bm{\theta}}) = \mathbbm{1}(\max{h_{\bm{\theta}}(\gamma(\bm{x}_u))>\tau}) \cdot H(\hat{\bm{y}}, h_{\bm{\theta}}(\Gamma(\bm{x}_u))),
\end{equation}
where $\Gamma$ denotes the strong augmentations that make the student learn in a harder way. 

To better suppress the erroneous pseudo labels, the mutual learning ~\cite{ke2020guided,fenga2020dmt} adopts two models $h_{{\bm{\theta}}_1}$ and $h_{{\bm{\theta}}_2}$ that are initialized differently and generate the pseudo labels $\hat{\bm{y}}_1$ and $\hat{\bm{y}}_2$ respectively. The same supervised loss as Equation~\ref{eq:supervised} is used whereas the unsupervised loss becomes:
\begin{equation}
\begin{multlined}
  \mathcal{L}_{u}({\bm{\theta}}_1,{\bm{\theta}}_2) = H(\hat{\bm{y}}_2, h_{{\bm{\theta}}_1}(\Gamma(\bm{x}_u))) \\
  + H(\hat{\bm{y}}_1, h_{{\bm{\theta}}_2}(\Gamma(\bm{x}_u))).
  \label{eq:mutual_loss}
\end{multlined}
\end{equation}
The incorrect pseudo labels are gradually filtered by the peer network~\cite{han2018co}, so both of the networks collaboratively improve during training.

\section{Robust Mutual Learning}
In this section, we propose robust mutual learning that suffers less from the coupling issue. Also, we involve pseudo label self-rectification to further enhance the supervision quality for unlabeled data.

\subsection{Techniques for Model Coupling Reduction}
\noindent\textbf{Indirect mutual learning with mean teachers.} Mutual learners possess different learning abilities due to different initialization and can online refine the pseudo labels from the counterpart. However, they may quickly converge to the same state and thereby suffer from the coupled noises. The model coupling in the early training phase will reduce the mutual learning to self-training. We conjecture that the coupling issue is caused by the direct interaction between the mutual learners. Therefore, we propose to transfer the knowledge in an indirect manner. Specifically, we generate pseudo labels from the mean teachers~\cite{tarvainen2017mean}, \ie, the exponential moving average (EMA) of the training models, so that each learner receives supervision from the peer mean teacher. Let $\tilde{h}_{{\bm{\theta}}_1}$ and $\tilde{h}_{{\bm{\theta}}_2}$ be the moving average model for $h_{{\bm{\theta}}_1}$ and $h_{{\bm{\theta}}_2}$ respectively, the one-hot mutual supervisions in Equation~\ref{eq:mutual_loss} become:
\begin{align}
  \hat{\bm{y}}_i = \onehot\left(\tilde{h}_{{\bm{\theta}}_i}(\gamma(\bm{x}_u))\right), i = 1,2.
\end{align}

Since each model learns from the temporal ensemble knowledge of the counterpart rather than its immediate state, the mutual learners
are kept diverged and reach a consensus more gradually 
\begin{figure}
  \begin{center}
  \includegraphics[width=0.48\textwidth]{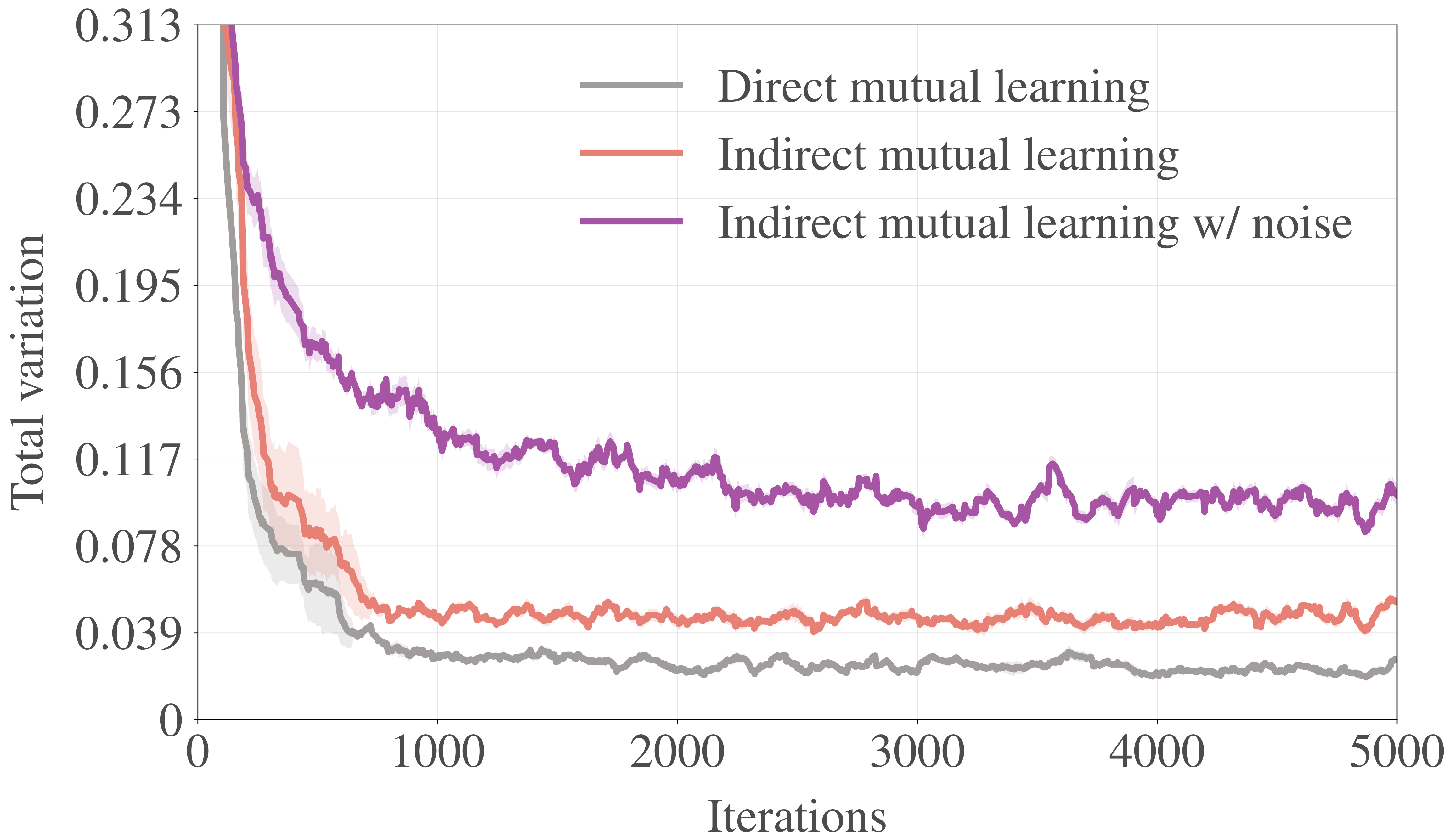}
  \end{center}
  \vspace{-1.6em}
  \caption{MNIST experiment.}
  \vspace{-1.4em}
  \label{fig:mnist}
 \end{figure}
compared to the direct mutual learning. To prove that indirect mutual learning can slow down the speed of model coupling, we train mutual networks of $3$ layers of MLPs on the MNIST dataset of which $1/60$ images are used as labeled data for the SSL setting. We measure the total variation distance of the softmax outputs of two teacher networks during training. Figure~\ref{fig:mnist} shows that the indirect teaching using mean teachers indeed leads to more divergent mutual learners whereas direct mutual learning causes model collapse (total variation distance approaches to zero) at the early training phase. 

In practice, each learner in our framework simultaneously learns from two supervisions (Figure~\ref{fig:diagram}d): the peer-supervision given by the peer mean teacher and self-supervision provided by its own mean teacher. We empirically find that learning from such an ensemble knowledge further improves the SSL performance. Therefore, the loss function for training two models is defined as
\begin{align}
\begin{split}
    \thinmuskip=1mu
    \thickmuskip=1mu
    \mathcal{L}_{u}({\bm{\theta}}_1,{\bm{\theta}}_2) &= H(\hat{\bm{y}}_2, h_{{\bm{\theta}}_1}(\Gamma(\bm{x}_u))) + H(\hat{\bm{y}}_1, h_{{\bm{\theta}}_1}(\Gamma(\bm{x}_u))), \\
     &\quad + H(\hat{\bm{y}}_1, h_{{\bm{\theta}}_2}(\Gamma(\bm{x}_u))) + H(\hat{\bm{y}}_2, h_{{\bm{\theta}}_2}(\Gamma(\bm{x}_u))).
\end{split}
     \label{eq:total_loss}
\end{align}

\vspace{0.4em} \noindent\textbf{Data augmentation and model noises.} Indirect mutual learning alleviates the model coupling to some extent but still requires different initialized mutual learners~\cite{ke2020guided,fenga2020dmt}. In contrast, we inject noises including data noise (\ie, data augmentation) and
model noises to perturb the mutual learners and ensure their divergency throughout the training process. The data augmentations applied for the student model include photometric augmentations in RandAugment~\cite{cubuk2020randaugment}.
On the other hand, the model noises, including dropout~\cite{srivastava2014dropout} and stochastic depth~\cite{huang2016deep}, are able to bring architectural perturbations. The benefit of the noise injection is two-fold. First, they force the student model to learn in a hard way, which has been proven crucially important for self-training~\cite{xie2020self}. Second, these perturbations prevent the mutual learners from closely tracking the state of the counterpart, hence the chance of model coupling is greatly reduced. 
In addition, we adopt CutMix~\cite{yun2019cutmix}, a kind of data augmentation that stitches training images based on a binary mask, to improve consistency learning.

In particular, we study the effectiveness of model noise using the MNIST experiment. As shown in Figure~\ref{fig:mnist}, we see clearly improved divergency of the mutual learners during training. Hence, the students consistently obtain complementary knowledge and collaboratively attain higher performance. 

\vspace{0.4em} \noindent\textbf{Heterogeneous architecture.} 
Last but not least, besides injecting model noises to bring architectural perturbations,
another straightforward way to ensure the architectural difference is to directly adopt distinct architectures. 
As opposed to using convolutional neural networks (CNN) with dissimilar backbones, we investigate Transformer~\cite{vaswani2017attention} which has shown tantalizing promise in the vision tasks. Since the Transformer is able to capture long-range dependency while CNN is good at modelling local context, it seems that a heterogeneous architecture with CNN and Transformer naturally fits the mutual learning framework as students learn complementary knowledge.
To validate our hypothesis, we experiment with the mutual learning between a CNN-based Deeplabv2 model and SETR~\cite{zheng2020rethinking}, a transformer-based segmentation network. The results in Section~\ref{sec:experiment} show that both networks, even without any noise injection, achieve significant gain from mutual learning. 

\subsection{Pseudo Label Self-rectification} 
Mutual learning heavily relies on the peer network but neglects the fact that the network itself can also identify suspicious pseudo labels. Since the pseudo labels are generated by the supervised model whose prediction confidence can be measured using the maximum softmax probability,  Monte-Carlo dropout or learned confidence estimates~\cite{devries2018leveraging,poggi2020uncertainty}, a natural idea is to cultivate the network's own ability to spot and suppress the unreliable pseudo labels. 

\vspace{0.4em} \noindent\textbf{Pseudo label correction.} To this end, we do not rely on the above uncertainty estimation techniques as they require a predefined threshold to remove the unreliable pseudo labels. Instead, we aim to online rectify the  pseudo labels that serve as progressively improved supervisions. Contrary to the peer-rectification, the self-rectification does not rely on any external knowledge. Such self-rectification, which stems  from~\cite{zhang2021prototypical}, rectifies the pseudo labels according to the \emph{class-wise confidence} that is estimated online. Formally, let $\bm{p}\in \mathbb{R}^{H\times W \times K}$ be the soft pseudo label for $\bm{x}_u$, \ie, $\bm{p} = h_{\bm{\theta}}(\bm{x}_u)$, whereas $\bm{p}_k\in \mathbb{R}^{H\times W}$ is the $k$th class probability given by the network output. We denote the initial state of this class probability as $\bm{p}_k^{0}$, which is dynamically re-weighted according to the class-wise confidence ${\bm\omega}_k \in \mathbb{R}^{H \times W \times K}$. Hence, the hard pseudo label $\tilde{\bm{y}}\in \mathbb{R}^{H\times W\times K}$ is online updated as:
\begin{equation}
  \tilde{\bm{y}} = \onehot\left(\{{\bm{p}}^t_0,{\bm{p}}^t_1,...,{\bm{p}}^t_K\}\right),\ \text{where}\ {\bm{p}}^t_k = {\bm\omega}_k \cdot \bm{p}_k^{0}.
  \label{eq:pseudo_denoising}
\end{equation}
Such rectified pseudo labels are further used as supervisions when optimizing Equation~\ref{eq:total_loss}, so that the students can always learn from denoised knowledge. 

\vspace{0.4em} \noindent\textbf{Class-wise confidence estimation.} Now the problem reduces to calculating the class-wise confidence $\bm{\omega_k}$. To achieve this, we model the clustered feature space as the exponential family mixture:
\begin{equation}
\begin{multlined}
    p(\bm{z}|\bm{\theta}) = \sum_{k=1}^{K} \pi_k p(\bm{z}|\bm{\theta}_k=\bm{\eta}_k) \\
    = \sum_{k=1}^{K} \pi_k \exp\left\{-d(\bm{z}, \bm{\eta}_k)\right\},
\end{multlined}
\end{equation}
where $z$ denotes the features of each pixel location and $\theta$ parameterizes this mixture model . The cluster center $\bm{\eta}_k\in \mathbb{R}^{C}$, or the \emph{prototype}, serves as the representative for the cluster, and $\pi_k$ is the weight for mixture distributions. Then, the posterior probability of the assignment $y$ for a given $\bm{z}$ can be computed as, 
\begin{equation}
    p(y=k|\bm{z}) = \frac{\pi_k \exp\left\{-d(\bm{z}, \bm{\eta}_k)\right\}}{\sum_{k'} \pi_{k'} \exp\left\{-d(\bm{z}, \bm{\eta}_{k'})\right\}},
    \label{eq:posterior}
\end{equation}
which is exactly the class-wise confidence useful to the pseudo label correction. When the  Euclidean distance is adopted, the class-wise confidence is essentially the softmax of distances to different prototypes, \ie:

\begin{equation}
  \bm{\omega}_k = \frac{ \pi_k \exp(-{\|\tilde{f}(\bm{x})-\bm{\eta}_k\|}) } { \sum_{k'} \pi_{k'} \exp(-{\|\tilde{f}(\bm{x})-\bm{\eta}_{k'}\|}) }.
\end{equation}
Here we use the feature space extracted by the mean teacher. Intuitively, the prediction confidence of belonging to $k$th class is measured by how much it is more close to $\bm{\eta}_k$ relative to other prototypes.

\vspace{0.4em} \noindent\textbf{Prototype calculation}
Ideally, prototypes should be computed using all the training images in each iteration, \ie,
\begin{equation}
\begin{aligned}
  \bm{\eta}_k = \frac{1}{|{\mathcal{D}_l\cup \mathcal{D}_u}|} \sum_{\bm{x}\in \{\mathcal{D}_l\cup \mathcal{D}_u\}}f_x \quad\quad\quad\quad\quad \\
  f_x = \mean_{i}^{HW} \left(\tilde{f}(\bm{x})^i \cdot \mathbbm{1}(\argmax(\tilde{h}(\bm{x})^i)==k)\right),
\end{aligned}
\end{equation}
where $i$ denotes the position index.
However, this is computationally prohibitive for the online training. As an approximation, we compute the moving average:
\begin{equation}
  \bm{\eta}_k \leftarrow \lambda \bm{\eta}_k + (1-\lambda) \bm{\eta}'_k,
\label{update}
\end{equation}
where $\bm{\eta}'_k$ is the feature mean of $k$th class for the current batch, and the momentum $\lambda$ is set to 0.9999. During training, this moving average estimation converges to the true cluster center as the features gradually evolve. Note that our approach is similar to clustering-based unsupervised learning~\cite{caron2018deep,caron2020unsupervised} that online computes the cluster centers, but we use the prototypes to rectify the soft pseudo label rather than cluster assignment for the following representation learning. Another benefit of using prototypes is that all the classes are treated equally regardless of their occurrence, so the segmentation suffers less from the class-imbalance issue.

The maintained $K$ prototypes serve as the network knowledge of the underlying feature clusters. The pseudo label rectification using such self-knowledge complements the denoising capability of the peer network, and both schemes can jointly refine the pseudo labels during training, leading to a significant performance boost for semi-supervised learning.  

\section{Experiments}
\label{sec:experiment}

\subsection{Dataset and Evaluation Protocol} 

\noindent\textbf{Datasets.} 
We conduct experiments on two commonly-used benchmarks for semantic segmentation.  

\begin{itemize}[leftmargin=*]
\item \emph{Cityscapes.} This is a finely annotated urban scene dataset with $19$ categories and consists of $2975$ training images. The original image size is $2048\times1024$. Following~\cite{french2020semi, hung2019adversarial}, we resize images to $1024\times512$ and randomly crop them to $512\times 256$ during training.
\item \emph{PASCAL VOC 2012.} The standard dataset contains $1464$ images that comprise $21$ classes including background. As a common practice, we also obtain $9118$ augmented images using~\cite{hung2019adversarial}. Images are randomly resized by $0.5\sim 1.5$ and cropped to $321\times 321$ for training.
\end{itemize}

\vspace{0.4em} \noindent\textbf{Evaluation protocol.} We evaluate the SSL performance under the learning from a varied amount of labeled images on two datasets. For the Cityscapes dataset, we randomly sample $1/30$, $1/8$ and $1/4$ images to construct the labeled data and regard the rest of the training images as unlabeled data. For PASCAL VOC, we use $1/100$, $1/50$, $1/20$ and $1/8$ images from the standard training set as labeled data while the remaining images in the standard training set together with the augmented images constitute the unlabeled data. For a fair comparison, we use the same data split as~\cite{french2019semi} when subsampling the labeled data. We measure the quantitative performance using mIoU, \ie, the mean of class-wise intersection over union.

\begin{table*}[htbp]
  \caption{Quantitative comparisons on Cityscapes dataset. We use a varied amount of labeled images and report the mIoU score (in percentage). Models are initialized with the pretrained weights on ImageNet except that CowMix~\cite{french2019semi} and DMT~\cite{fenga2020dmt} report the performance using the COCO initialization.}
  \label{cityscapes-table}
  \centering
  \small
  \begin{tabular}{ccccc}
    \toprule
    Cityscapes & \multicolumn{4}{c}{\# Labels} \\
    \midrule
    Methods (ImageNet init.) & 1/30 (100) & 1/8 (372) & 1/4 (744) & Full (2975) \\
    \midrule
    Deeplabv2~\cite{chen2017deeplab} & - & 59.3 & 61.9 & 66.0 \\
    AdvSSL~\cite{hung2019adversarial} & - & 57.1 & 60.5 & 66.2 \\
    S4GAN~\cite{mittal2019semi} & - & 59.3 & 61.9 & 65.8 \\
    CutMix~\cite{french2020semi} & 51.20$\scriptstyle\pm$\tiny2.29 & 60.34$\scriptstyle\pm$\tiny1.24 & 63.87$\scriptstyle\pm$\tiny0.71 & 67.68$\scriptstyle\pm$\tiny0.37 \\
    CowMix$^\ast$~\cite{french2019semi} & 49.01$\scriptstyle\pm$\tiny2.58 & 60.53$\scriptstyle\pm$\tiny0.29 & 64.10$\scriptstyle\pm$\tiny0.82 & - \\
    ClassMix~\cite{olsson2021classmix} & 54.07$\scriptstyle\pm$\tiny1.61 & 61.35$\scriptstyle\pm$\tiny0.62 & 63.63$\scriptstyle\pm$\tiny0.33 & - \\
    ECS~\cite{mendel2020semi} & - & 60.26$\scriptstyle\pm$\tiny0.84 & 63.77$\scriptstyle\pm$\tiny0.65 & - \\
    DMT$^\ast$~\cite{fenga2020dmt} & 54.8 & 63.0 & - & 68.2 \\
    \midrule
    Deeplabv2 (reimplement) & 45.40$\scriptstyle\pm$\tiny1.04 & 56.08$\scriptstyle\pm$\tiny0.27 & 60.91$\scriptstyle\pm$\tiny0.62 & 67.14$\scriptstyle\pm$\tiny0.32 \\
    \emph{Ours} &  \textbf{58.25}$\scriptstyle\pm$\tiny1.36 &  \textbf{63.97}$\scriptstyle\pm$\tiny0.46 &  \textbf{66.15}{$\scriptstyle\pm$\tiny0.06} &  -\\
    \bottomrule
  \end{tabular}
\end{table*}%

\begin{figure*}[t]
  \center
  \small
  \setlength\tabcolsep{0pt}
  {
  \renewcommand{\arraystretch}{0.6}
  \begin{tabular}{@{}cccc@{}}
      
      \includegraphics[width=0.5\columnwidth]{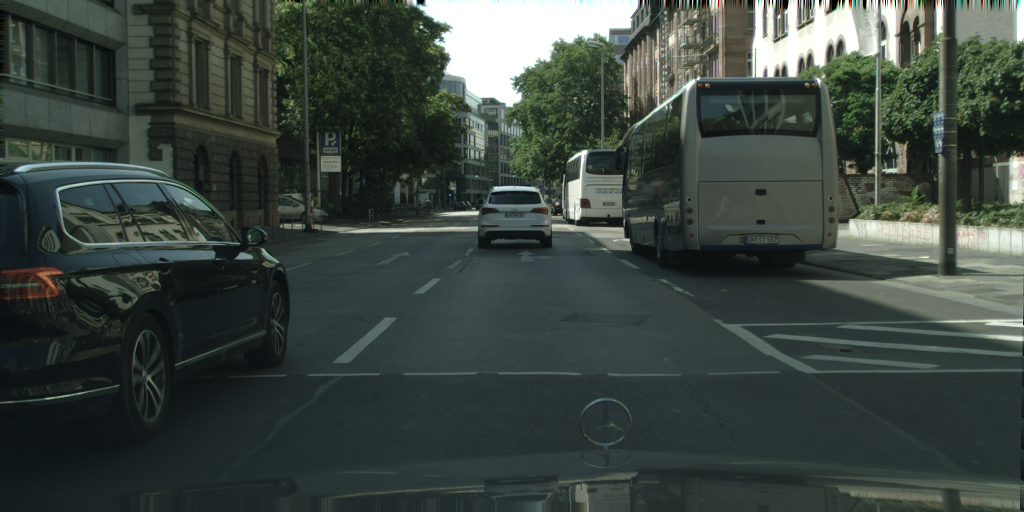}&
      \includegraphics[width=0.5\columnwidth]{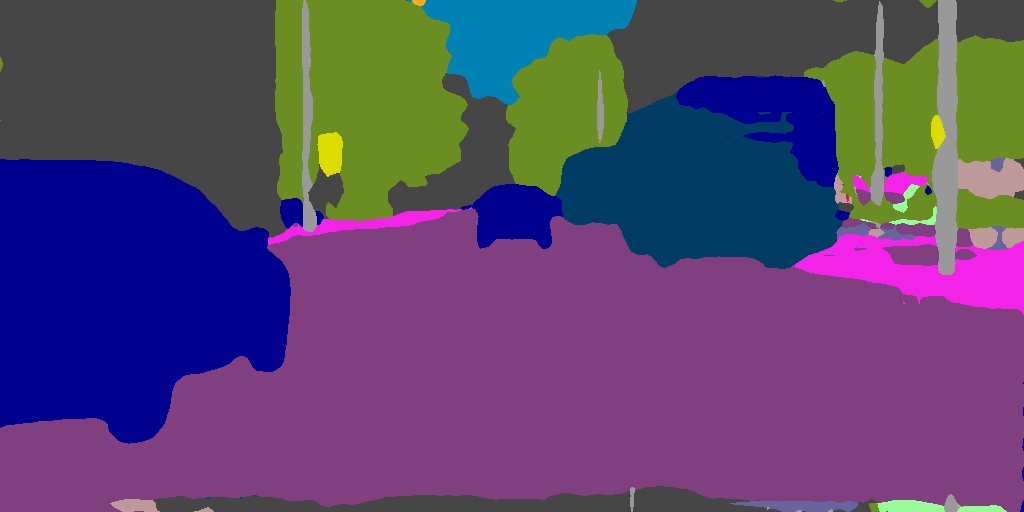}&
      \includegraphics[width=0.5\columnwidth]{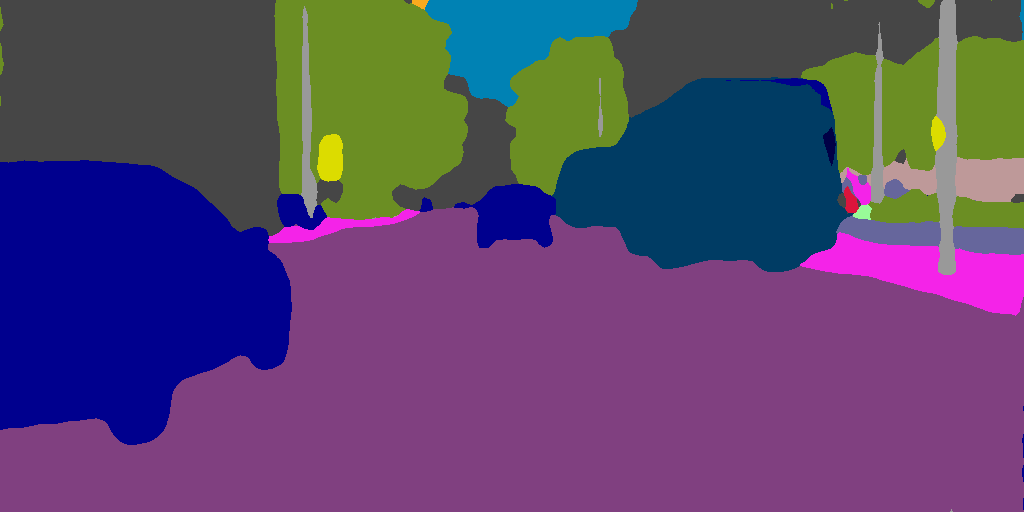}&
      \includegraphics[width=0.5\columnwidth]{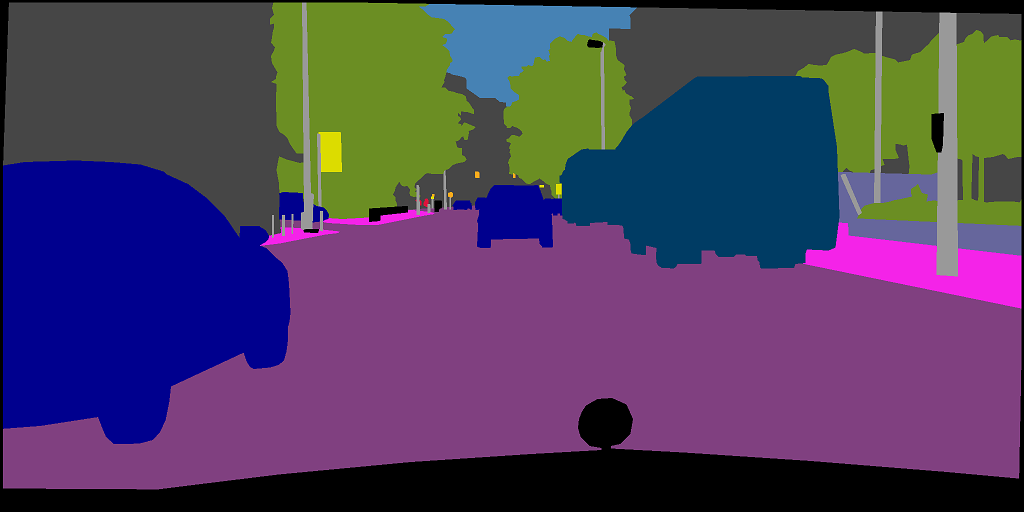}\\
  
      \includegraphics[width=0.5\columnwidth]{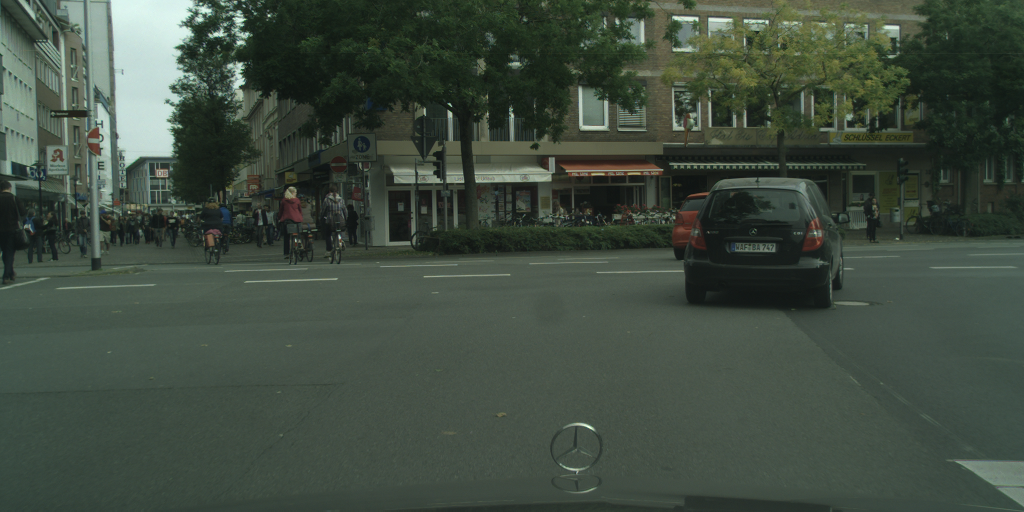}&
      \includegraphics[width=0.5\columnwidth]{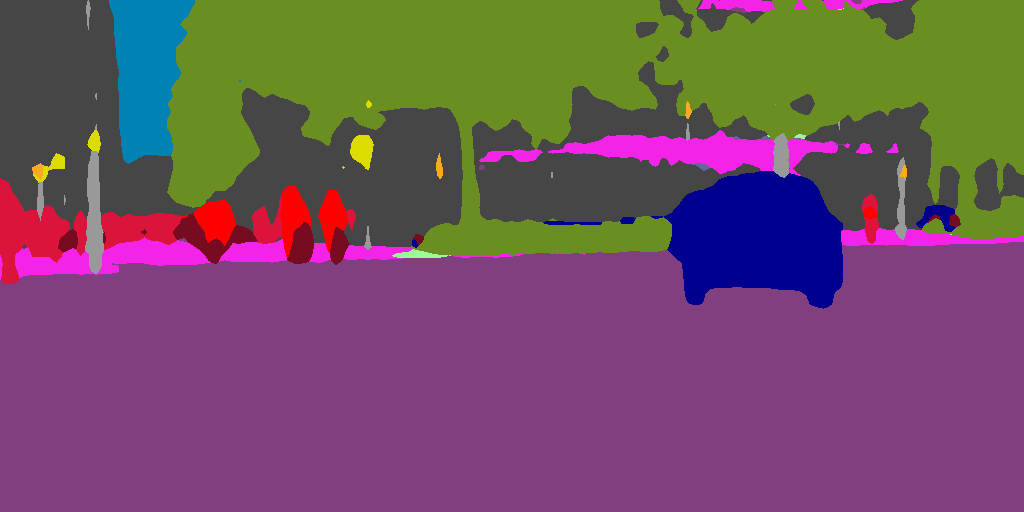}&
      \includegraphics[width=0.5\columnwidth]{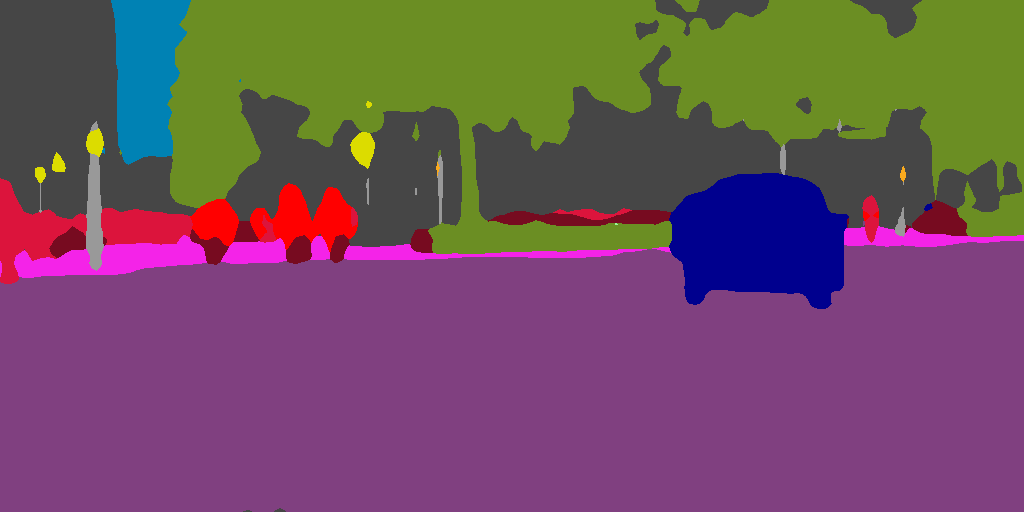}&
      \includegraphics[width=0.5\columnwidth]{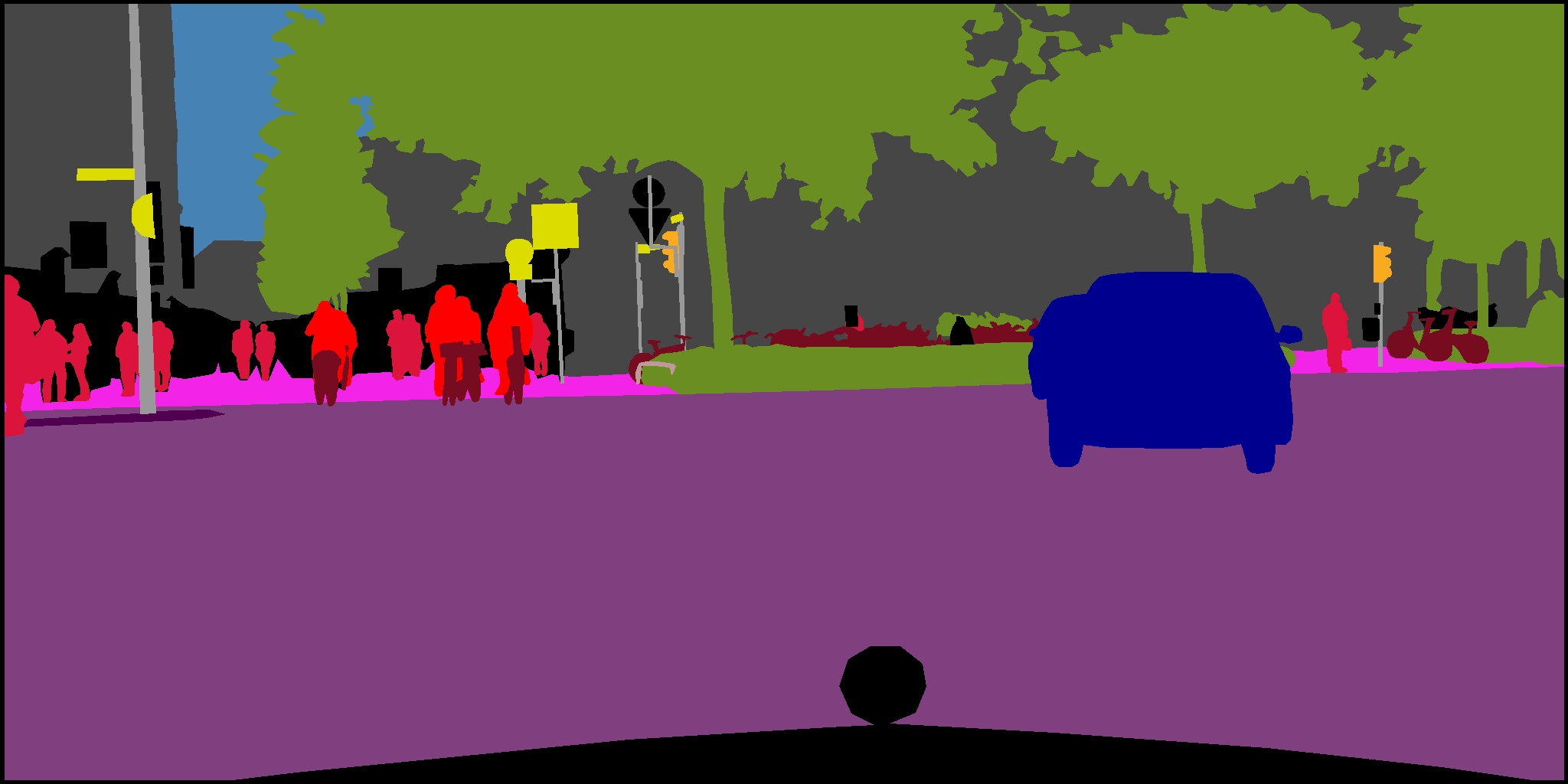}\\

      \includegraphics[width=0.5\columnwidth]{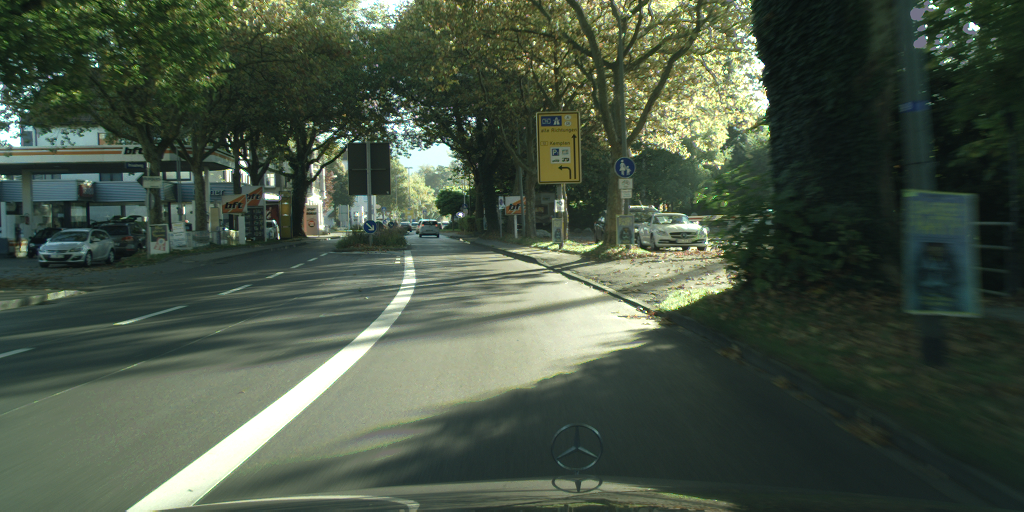}&
      \includegraphics[width=0.5\columnwidth]{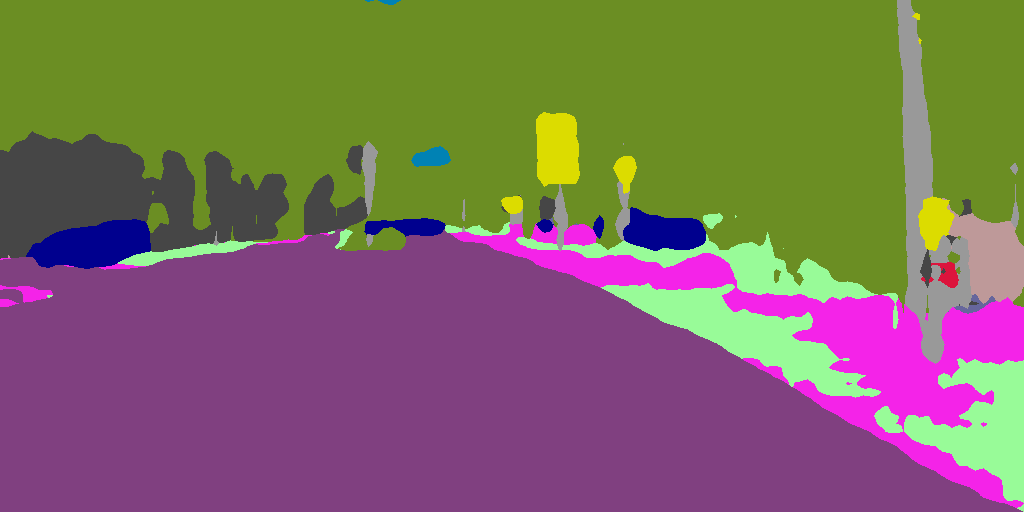}&
      \includegraphics[width=0.5\columnwidth]{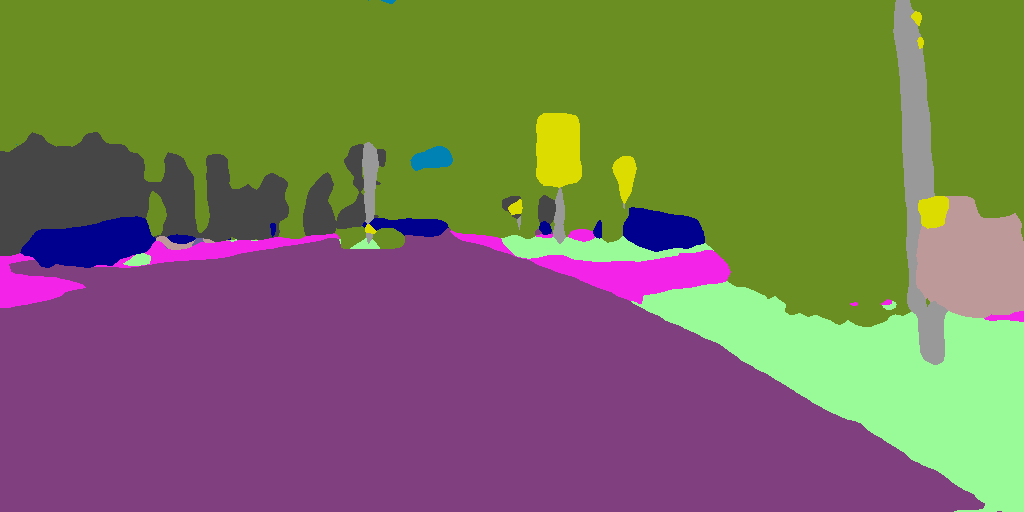}&
      \includegraphics[width=0.5\columnwidth]{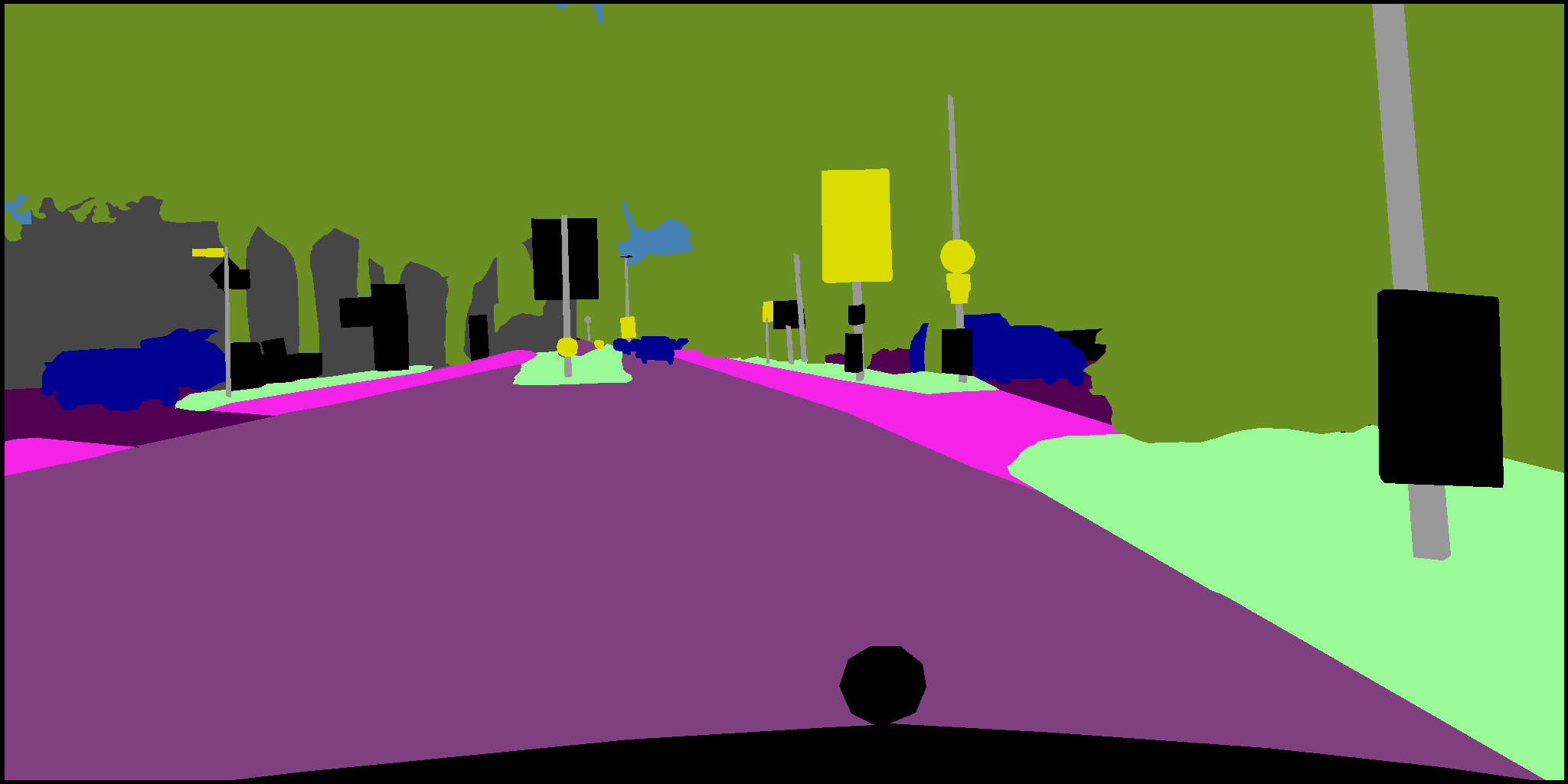}\\
      
      Image & CutMix~\cite{french2020semi} & Robust mutual learning & Ground truth\\
  \end{tabular}
  }
  \vspace{-0.5em}
  \caption{Qualitative comparison on Cityscapes dataset. Models are trained using 1/8 training images as labeled data.}
  \vspace{-1.0em}
  \label{fig:qualitative}
  \end{figure*}
  
\vspace{0.5em}
\subsection{Implementation Details} \label{Implementation}
\vspace{0.5em}
In our experiments, the CNN-based segmentation network adopts the Deeplabv2~\cite{chen2017deeplab} architecture
with the ResNet101~\cite{he2016deep} as the backbone. We apply photometric augmentations within RandAugment~\cite{cubuk2020randaugment} as well as CutMix~\cite{yun2019cutmix} for data augmentation. In terms of the model noises, we introduce dropout~\cite{srivastava2014dropout} only for the last FC layer with the dropout rate as $0.5$, and apply stochastic depth~\cite{huang2016deep} to the residual blocks of backbone with layer-wise survival probability as $0.8$. The proposed framework is optimized with SGD solver. The learning rate is $1\mathrm{e}{-4}$ with polynomial decay. We adopt \emph{stage-wise training strategy}, \ie, once the mutual learning converges we recompute the pseudo labels and then start the next training stage. In our experiments, we observe performance saturation after training two stages. For Cityscapes dataset, we train $90$k iterations for each stage with the batchsize of $4$; whereas each training stage of PASCAL VOC lasts $60$k iterations with the batchsize of $10$. For both datasets, we initialize the backbone with ImageNet~\cite{deng2009imagenet} pre-trained weights. Besides, we also use the pre-trained weights from the MS COCO dataset~\cite{lin2014microsoft} to initialize the model specifically for PASCAL VOC as suggested by ~\cite{hung2018adversarial,ke2020three}. The training takes around $1.5$ days using $4$ Tesla V100 GPUs. Our method is implemented with Pytorch~\cite{paszke2019pytorch} and we plan to make the code publicly available.

\begin{table*}[htbp]
  \caption{Quantitative comparisons on PASCAL VOC dataset. We use a varied amount of labeled images and report the mIoU scores (in percentage). ImageNet initialization is used.}
  \label{pascal-imagenet-table}
  \centering
  \small
  \begin{tabular}{cccccc}
    \toprule
    PASCAL VOC & \multicolumn{5}{c}{\# Labels} \\
    \midrule
    Methods (ImageNet init.) & 1/100 & 1/50 & 1/20 & 1/8 & Full (10582) \\
    \midrule
    Deeplabv2~\cite{chen2017deeplab} & - & 48.3 & 56.8 & 62.0 & 70.7 \\
    AdvSSL~\cite{hung2019adversarial} & - & 49.2 & 59.1 & 64.3 & 71.4 \\
    S4GAN~\cite{mittal2019semi} & - & 60.4 & 62.9 & 67.3 & 73.2 \\
    ICT~\cite{verma2019interpolation} & 35.82 & 46.28 & 53.17 & 59.63 & 71.50 \\
    CutMix~\cite{french2020semi} & \textbf{53.79} & \textbf{64.81} & 66.48 & 67.60 & 72.54 \\
    \midrule
    Deeplabv2 (reimplemtent) & 36.25 & 46.30 & 55.92 & 63.73 & 72.62 \\
    \emph{Ours} & 53.77 & 64.63 & \textbf{68.82} & \textbf{70.34} & -\\
    \bottomrule
  \end{tabular}
  \vspace{-0.8em}
\end{table*}%

\begin{table*}[htbp]
  \caption{Quantitative comparisons on PASCAL VOC dataset. We use a varied amount of labeled images and report the mIoU score (in percentage). COCO initialization is used.}
  \label{pascal-coco-table}
  \centering
  \small
  \begin{tabular}{cccccc}
    \toprule
    PASCAL VOC & \multicolumn{5}{c}{\# Labels } \\
    \midrule
    Methods (COCO init.) & 1/100 & 1/50 & 1/20 & 1/8 & Full (10582) \\
    \midrule
    Deeplabv2~\cite{chen2017deeplab} & - & 53.2 & 58.7 & 65.2 & 73.6 \\
    AdvSSL~\cite{hung2019adversarial} & - & 57.2 & 64.7 & 69.5 & 74.9 \\
    S4GAN~\cite{mittal2019semi} & - & 60.9 & 66.4 & 69.5 & 73.9 \\
    DMT~\cite{fenga2020dmt} & 63.04 & 67.15 & 69.92 & 72.70 & 74.75 \\
    ECS~\cite{mendel2020semi} & - & - & - & 72.95 & - \\
    ClassMix~\cite{olsson2021classmix} & 54.18 & 66.15 & 67.77 & 71.00 & - \\
    \midrule
    \emph{Ours} & \textbf{63.60} & \textbf{68.66} & \textbf{71.19} & \textbf{73.38} & - \\
    \bottomrule
  \end{tabular}
  \vspace{-1.0em}
\end{table*}%

\subsection{Comparisons with Previous Works} 

We comprehensively compare our method with state-of-the-art methods which can be categories as: 1) adversarial based methods, \ie, AdvSSL~\cite{hung2019adversarial} and S4GAN~\cite{mittal2019semi}; 2) methods that encourage consistency, \ie, ICT~\cite{verma2019interpolation}, CutMix~\cite{french2020semi}, CowMix~\cite{french2019semi} and ClassMix~\cite{olsson2021classmix}; 3) two recent works that also collaboratively train networks, \ie, ECS~\cite{mendel2020semi} and DMT~\cite{fenga2020dmt}. For fair comparisons, we report the performance reported by their original papers under the same SSL setting. 

Table~\ref{cityscapes-table} shows the quantitative comparison on Cityscapes dataset. Our method outperforms the prior leading approach significantly. We can see that the performance advantage becomes more evident when using fewer labeled images for training. In particular, when using $1/30$ labeled data, our method improves the mIoU by $3.45\%$ over prior best approach. Notably, our training on $1/4$ labeled images gives mIoU on par with supervised learning, proving the data efficiency of our approach. Besides, comparing with DMT that also adopts mutual learning, our robust mutual learning is more powerful to denoise the pseudo labels and hence yields higher performance. We visualize the qualitative results in Figure~\ref{fig:qualitative}. Compared with CutMix, our method excels at discriminating tiny objects and resolving ambiguity, giving results more consistent with the ground truth.

When conducting experiments on PASCAL VOC, we compare different methods using two types of initializations. Table~\ref{pascal-imagenet-table} shows the performance using the ImageNet initialization. We improve the mIoU by $2.34\%$ and $2.74\%$ for $1/20$ and $1/8$ labeled images respectively. For lower data regime, our method achieves almost the same performance as the prior leading approach. On the other hand, the COCO dataset contains images with more similar content as the PASCAL, so the COCO initialization generally leads to better SSL performance. As shown in Table~\ref{pascal-coco-table}, our method yields superior mIoU scores in all the data regimes.

\subsection{Discussions}

\vspace{0.4em} \noindent\textbf{High divergence throughout the training process.} Figure~\ref{fig:mnist} shows the proposed method maintains high divergence throughout the training process. In the early training phase, the pseudo labels are extremely noisy, so we need to maintain sufficiently divergent mutual learners to combat the noises in the pseudo labels generated by each model. Towards the end of training, the mutual learners may reach to a consensus on easy dataset. But real data lies on a more complicated manifold, so the model with limited capacity cannot fit the whole dataset perfectly and inevitably make mistakes on some hard samples. Since the divergence is measured using the whole training samples, a high divergence does not necessarily mean that the models always give contradictory predictions for individual samples. Rather, this can be understood as that models do not make mistakes on the same type of data: one model may be good at predicting 90\% samples whereas another may be good at another 90\% samples.

\vspace{0.4em} \noindent\textbf{Ablation study.} We quantify the effectiveness of each proposed component in Table~\ref{table:ablation}. We see that indirect mutual learning (IML) improves the mIoU of supervised baseline by $4.06\%$. The model noises and inputs noises are essential to reduce the mutual coupling and can jointly contribute to the mIoU gain by $1.16\%$. Moreover, with the pseudo label self-rectification, the single-stage robust mutual learning (RML) obtains mIoU improvement by $1.6\%$, and the second training stage achieves even higher performance. Overall, the improvement of our two-stage RML is substantial with the mIoU gain over the supervised baseline by $7.62\%$. 

\begin{table}[htbp]
  \small
  \centering 
  \caption{Ablation study in terms of mIoU. We ablate proposed components on Cityscapes 1/8 labels. IML: indirect mutual learning. RML: robust mutual learning.}
  \begin{tabular}{cc}
  \toprule
  Components & mIoU \\ 
  \midrule
  \makecell[c]{Supervised baseline} & 56.03 \\ 
  \midrule
  IML & 60.09\\
  \makecell[c]{IML (+model noises)} & 60.51 \\
  \makecell[c]{IML (+input noises)} & 60.49 \\
  \makecell[c]{IML (+full noises)} & 61.25 \\
  RML &  62.85 \\
  \makecell[c]{RML (two stages)} &\textbf{63.65} \\
  \bottomrule
  \end{tabular}
  \label{table:ablation}
\end{table}%

\begin{figure}[!t]
  \small
  \begin{center}
  \includegraphics[width=0.44\textwidth]{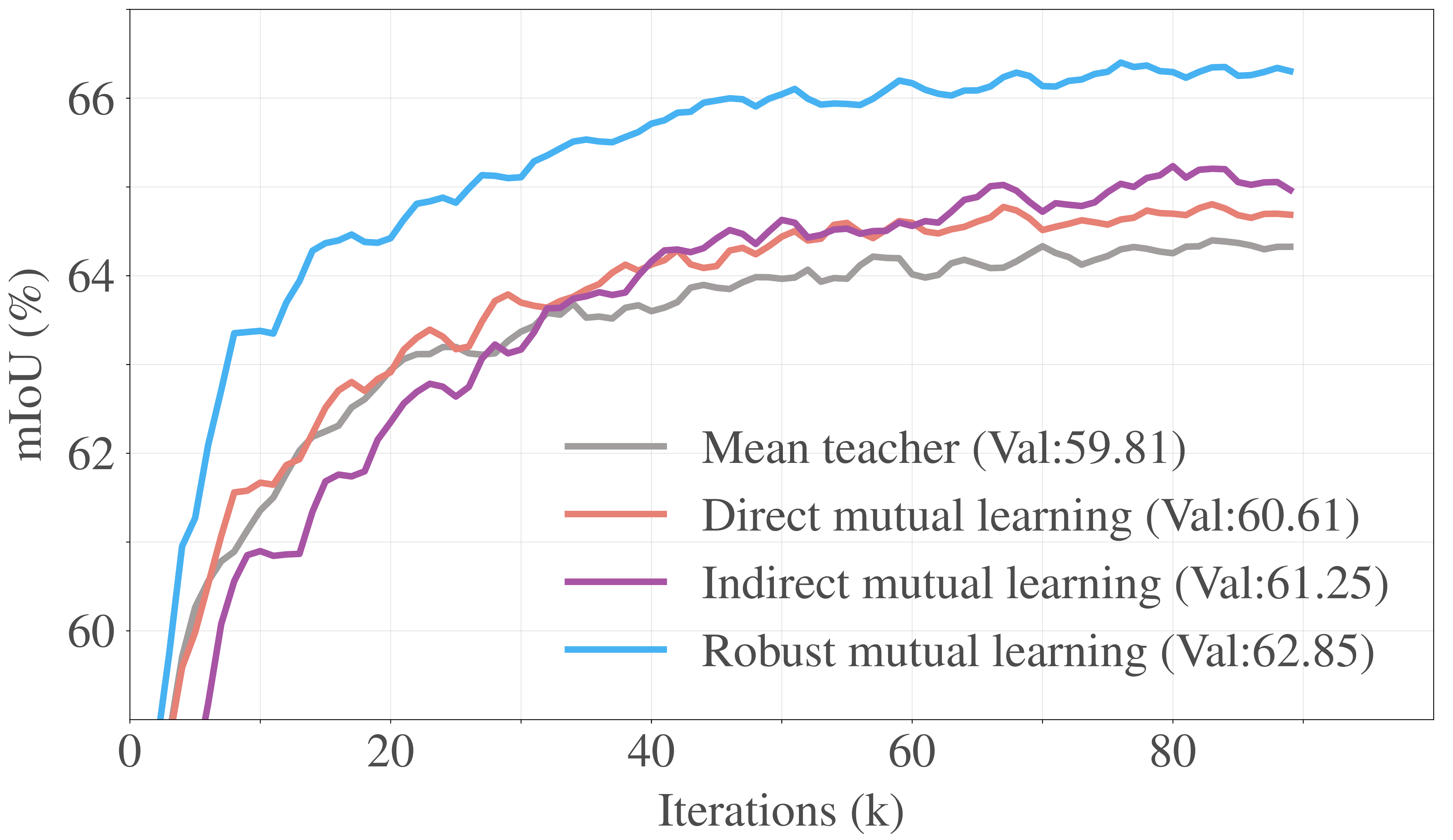}
  \end{center}
  \vspace{-1.8em}
  \caption{The process of pseudo label online refinement during the first stage training. The experiments are conducted on Cityscapes dataset using Deeplab v2 backbone.}
  \vspace{-1.8em}
  \label{fig:pseudo_label}
 \end{figure}

\vspace{0.4em} \noindent\textbf{The online refinement of pseudo labels.} The key motivation of our robust mutual learning is to online refine the pseudo labels so that better supervisions can be obtained for self-training. Figure~\ref{fig:pseudo_label} visualizes how the pseudo labels progress in the course of training. Comparing with the MeanTeacher~\cite{tarvainen2017mean}, direct mutual learning produces better pseudo labels as noises can be gradually suppressed by the peer network. In comparison, indirect mutual learning reduces the mutual coupling and thus leads to more effective co-teaching. Finally, much cleaner pseudo labels are produced in the robust mutual learning thanks to the strong denoising capability of the self-rectification scheme. 
 
 \vspace{0.4em} \noindent\textbf{Heterogeneous architecture.} To validate the hypothesis that heterogeneous architecture is effective to reduce the model coupling, we adopt CNN and Transformer network to the mutual learning framework. Specifically, the student models adopt Deeplabv2 and SETR~\cite{zheng2020rethinking} architecture respectively. We study such hybrid architecture without noise injection nor self-rectification. Table~\ref{table:component-table2} gives the result and we highlight several interesting findings. First, comparing to the baseline, all the architectural combinations benefit largely from mutual learning, proving the universality of the method. Second, the mutual learning between Deeplabv2 and SETR  demonstrates the best performance. Notably, Deeplabv2 improves the mIoU by $2.71\%$ when mutually trained with SETR compared to the mutual training with the same architecture. On the other hand, instead of being 
pulled down by the Deeplabv2, Transformer-based SETR obtains useful peer-supervision and achieves the highest performance in such hybrid mutual learning. 

\vspace{0.4em} \noindent\textbf{Sensitivity analysis.} One advantage of this work is that our method involves only a few hyperparameters. Table~\ref{table:sensitivity} gives sensitivity analysis on $\lambda$ in Equation~\ref{update}, from which we can see that the performance is insensitive to $\lambda$. More sensitivity analysis can be found in the supplementary material.

\begin{table}[!tb]
  \caption{Heterogeneous architecture (\eg, CNN vs Transformer) fits the mutual learning framework due to less model coupling (results on Cityscapes using $1/8$ labeled data).}
  \label{table:component-table2}
  \centering
  \small
  \begin{tabular}{cc}
    \toprule
     Architecture & mIoU\\
     \midrule
     Deeplabv2 baseline & 55.86\\
     SETR~\cite{zheng2020rethinking} baseline & 59.77\\
    \midrule
    Deeplabv2-Deeplabv2 IML & 61.25 / 61.53 \\
    SETR-SETR IML & 64.15 / 63.74 \\
    Deeplabv2-SETR IML & 63.96 / 64.35 \\
    \bottomrule
  \end{tabular}
\end{table}

\begin{table}[!tb]
  \caption{Sensitivity to $\lambda$ using 1/8 labeled data on Cityscapes dataset for the first training stage.}
  \label{table:sensitivity}
  \centering
  \small
  \begin{tabular}{c|ccc}
    \toprule
     1-$\lambda$ & 1e-3 & 1e-4 & 1e-5\\
     \midrule
    mIoU & 62.67 & 62.85 & 62.69 \\
    \bottomrule
  \end{tabular}
\end{table}

\vspace{0.4em} \noindent\textbf{Limitation.} \label{Limitation}
While our method has shown state-of-the-art performance, the performance advantage on PASCAL VOC dataset is not as obvious as Cityscapes. We conjecture that this is because PASCAL VOC dataset is not finely annotated where the background class is not clean. Thus, the prototypes computed for such class may not be accurate. Besides, it takes many iterations to update the class-wise prototypes since each image in this dataset only contains $2\sim3$ classes. We believe there is still room for further improvement along with our RML, which we leave for future work.

\section{Conclusion}
In this paper, we identify the model coupling issue in the current mutual learning works and comprehensively explore factors including aggressive input augmentation, model perturbation and adopting heterogeneous architectures for the mutual learners to alleviate the issue.
To enhance the accuracy of peer supervision, we propose a self-rectification scheme that explicitly refines the pseudo labels by leveraging the self-knowledge of the network.
The resulting framework, termed robust mutual learning, could generate better pseudo labels when simultaneously conducting the self-training and mutual learning. Our approach is a step further to narrow the gap with supervised learning for low-data regime and hopefully will inspire more research on denoising pseudo labels for effective semi-supervised learning.

\clearpage

{\small
\bibliographystyle{ieee_fullname}
\bibliography{egbib}
}

\end{document}


\title{Robust Mutual Learning for Semi-supervised Semantic Segmentation (Supplementary Material)}

\author{First Author\\
Institution1\\
Institution1 address\\
{\tt\small firstauthor@i1.org}
\and
Second Author\\
Institution2\\
First line of institution2 address\\
{\tt\small secondauthor@i2.org}
}
\maketitle

We provide more implementation details in Section~\ref{details}, including architecture, data augmentation and optimization, as well as more analytical experiments in Section~\ref{analytical}, including tiny classes improvement and hyperparameters analysis.

\section { More Implementation Details} \label{details}
In our experiments, the CNN-based segmentation network adopts the Deeplabv2~\cite{chen2017deeplab} architecture with the ResNet101~\cite{he2016deep} as the backbone. A slight difference is that after the ASPP module, we introduce an additional $3\times 3$ convolution layer for reducing the number of channels and a $1\times 1$ convolution layer to obtain the pixel-wise features for prototype calculation. The same changes can be found in ~\cite{zheng2019unsupervised, zhang2021prototypical}. For transformer, we follow SETR~\cite{zheng2020rethinking} and use SETR-Na\"ive-Base model.
To train the transformer, we down sample Cityscapes images to $1024\times512$ and randomly crop them to $512\times256$. The word length is $16\times16$, and 12 heads are adopted. The optimizer is AdamW with learning rate set as $3e-4$. Step learning rate scheduler is adopted and weight decay is set as $1e-3$. Algorithm~\ref{alg} provides more training detail.

\vspace{-0.3em}
\SetKwComment{Comment}{$\triangleright$\ }{} \label{alg}
\SetCommentSty{mycommfont}
\SetKwInput{KwInput}{Input}
\SetKwInput{KwOutput}{Output}
\begin{algorithm}[tbh]
\small
\DontPrintSemicolon
\KwInput{Labeled samples: $\{(x_i,y_i) | i\in L\}$; Unlabeled samples: $\{x_u | u\in U\}$; the baseline model: $h_{\bm{\theta}}$;}
Generate soft pseudo label: $p_{0} \leftarrow h_{\bm{\theta}}(U)$;\;
Prototype initialization: $\eta_c^1, \eta_c^2 \leftarrow (h_{\bm{\theta}},L \cup U)$;\quad
Model initialization: $h_{\bm{\theta_1}}, h_{\bm{\theta_2}}, \tilde{h}_{\bm{\theta_1}}, \tilde{h}_{\bm{\theta}_2} \leftarrow h_{\bm{\theta}}$;\;
\For{$m\gets0$ \KwTo \text{epochs}}{
      Sample labeled images $x_i, y_i$;\Comment*[r]{Labeled training}
      Update model $h_{\bm{\theta_1}}$ and $h_{\bm{\theta_2}}$ with $(x_i,y_i)$;\;
      \;
      Sample two unlabeled images $x_{u1}, x_{u2}$;\Comment*[r]{Unlabeled training}
      Random generate rectangle mask $m$, following ~\cite{french2020semi};\;
      $x_u \leftarrow m * x_{u1} + (1-m) * x_{u2}$;\Comment*[r]{CutMix of unlabeled images}
      $\hat{y}_u^1 \leftarrow \textnormal{MixRectify}(\tilde{h}_{\theta_1}, x_{u1}, x_{u2}, m, \eta_c^1)$;\;
      $\hat{y}_u^2 \leftarrow \textnormal{MixRectify}(\tilde{h}_{\theta_2}, x_{u1}, x_{u2}, m, \eta_c^2)$;\Comment*[r]{Self-rectification}
      Update model $h_{\theta_1}$ and $h_\theta_2$ with $(x_u,\hat{y}_u^1)$ and $(x_u,\hat{y}_u^2)$;\Comment*[r]{Mutual learning}
      \;
      Calculate the batch prototype $\eta_c^{1'}, \eta_c^{2'}$ ;\Comment*[r]{Prototypes of current batch}
      $\eta_c^1 \leftarrow \lambda \eta_c^1 + (1-\lambda) \eta_c^{1'}$;\quad $\eta_c^2 \leftarrow \lambda \eta_c^2 + (1-\lambda) \eta_c^{2'}$;\Comment*[r]{Update prototypes}
      Update the EMA model $\tilde{h}_{\bm{\theta_1}}$, $\tilde{h}_{\bm{\theta_2}}$;\Comment*[r]{Update teacher models}
}
  \;
\SetKwProg{Fn}{Function}{:}{}
  \Fn{\textnormal{MixRectify}($f_{\bm{\theta}}$, $x_1$, $x_2$, $m$, $\eta_c$)}{
        $\hat{y}_1 \leftarrow \textnormal{Denoise}(f_{\bm{\theta}}, x_1, \eta_c)$;\quad
        $\hat{y}_2 \leftarrow \textnormal{Denoise}(f_{\bm{\theta}}, x_2, \eta_c)$;\;
        $\hat{y} \leftarrow m * \hat{y}_1 + (1-m) * \hat{y}_2$;\Comment*[r]{CutMix of rectified pseudo labels}
        \KwRet $\hat{y}$\;
  }

\SetKwProg{Fn}{Function}{:}{}
  \Fn{\textnormal{Denoise}($f_{\bm{\theta}}$, $x$, $\eta_c$)}{
        Calculate the denoising weight $\gamma^{(i,k)}$, then update the pseudo label $\hat{y}^{(i,k)}$;\;
        \KwRet $\hat{y}^{(i,k)}$\;
  }

\caption{Robust Mutual Learning }
\label{RML}
\end{algorithm}

\clearpage

\section{More Analytical Experiments} \label{analytical}
\paragraph{Tiny classes benefit more.}
We show the performance comparison per  class in
Table~\ref{table:perclass}. The proposed robust mutual learning gets significant improvement on some tiny or small classes, such as fence, truck and motor.
This probably attributes to the powerful prototypes that treat every class equally and thus could handle imbalance problems in semantic segmentation.
Figure~\ref{fig:qualitative1} gives more visual comparison results. Compared with CutMix, our method excels at discriminating tiny objects as well as resolving ambiguity, giving more consistent results with the ground truth.

\begin{table*}[h!]
  \centering
  \footnotesize
  \caption{Per-class IoU comparison on Cityscapes with 1/8 labeled images. }
  \setlength\tabcolsep{1pt}{
  \begin{tabular}{c|*{19}{c}|c}
  \toprule
  & \multicolumn{1}{c}{\begin{sideways}road\end{sideways}} & \multicolumn{1}{c}{\begin{sideways}sideway\end{sideways}} & \multicolumn{1}{c}{\begin{sideways}building\end{sideways}} & \multicolumn{1}{c}{\begin{sideways}wall\end{sideways}} & \multicolumn{1}{c}{\begin{sideways}fence\end{sideways}} & \multicolumn{1}{c}{\begin{sideways}pole\end{sideways}} & \multicolumn{1}{c}{\begin{sideways}light\end{sideways}} & \multicolumn{1}{c}{\begin{sideways}sign\end{sideways}} & \multicolumn{1}{c}{\begin{sideways}vege.\end{sideways}} & \multicolumn{1}{c}{\begin{sideways}terrace\end{sideways}} & \multicolumn{1}{c}{\begin{sideways}sky\end{sideways}} & \multicolumn{1}{c}{\begin{sideways}person\end{sideways}} & \multicolumn{1}{c}{\begin{sideways}rider\end{sideways}} & \multicolumn{1}{c}{\begin{sideways}car\end{sideways}} & \multicolumn{1}{c}{\begin{sideways}truck\end{sideways}} & \multicolumn{1}{c}{\begin{sideways}bus\end{sideways}} & \multicolumn{1}{c}{\begin{sideways}train\end{sideways}} & \multicolumn{1}{c}{\begin{sideways}motor\end{sideways}} & \multicolumn{1}{c}{\begin{sideways}bike\end{sideways}} & \multicolumn{1}{|l}{mIoU}\\
  \midrule
  \makecell[c]{Supervised \\baseline} & 96.09 & 70
  14& 85.77 & 34.22 &24.96&38.59&35.68&50.43&87.23&50.55&89.90&62.72&38.17&87.76&32.39&51.99&38.56&38.15&60.17&56.03 \\
  \midrule
  stage 1 & 96.91 & 75.56 &87.90&43.28&35.45&39.95&41.75&55.69&88.47&51.49&90.74&68.84&46.91&90.26&46.38&71.94&52.03&46.38&64.22&62.85\\
  \midrule
  stage 2 &97.15&77.33&88.48&41.17&41.52&41.96&44.76&58.70&89.29&54.39&91.89&70.46&48.26&90.82&44.56&64.09&46.49&51.98&65.92&63.65\\
  \bottomrule
  \end{tabular}}
  \label{table:perclass}
\end{table*}

\begin{figure*}[t]
  \center
  \small
  \setlength\tabcolsep{0pt}
  {
  \renewcommand{\arraystretch}{0.6}
  \begin{tabular}{@{}cccc@{}}
      
      \includegraphics[width=0.25\columnwidth]{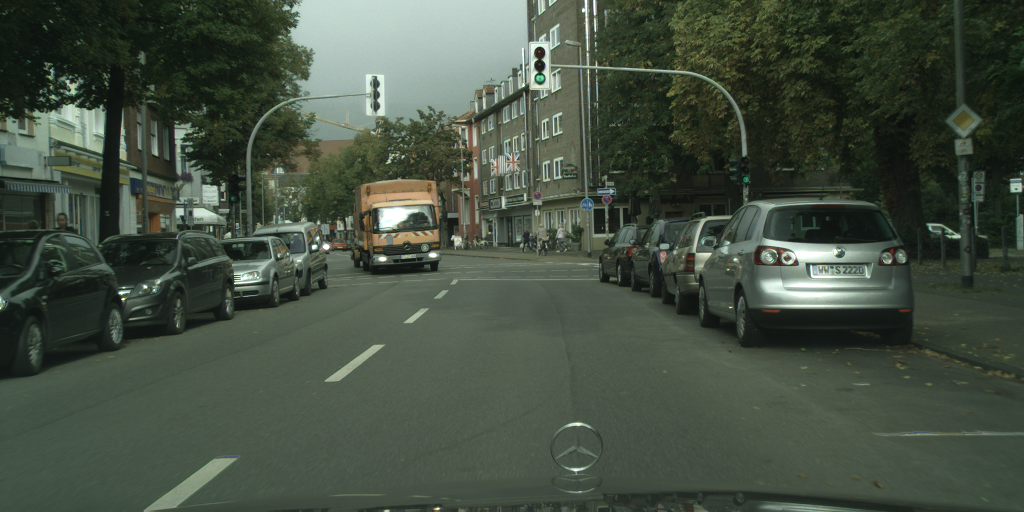}&
      \includegraphics[width=0.25\columnwidth]{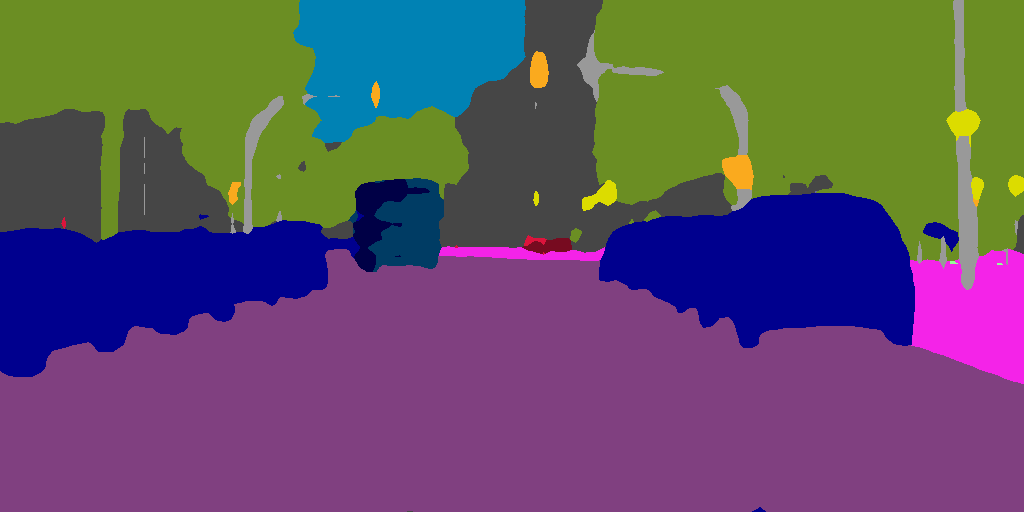}&
      \includegraphics[width=0.25\columnwidth]{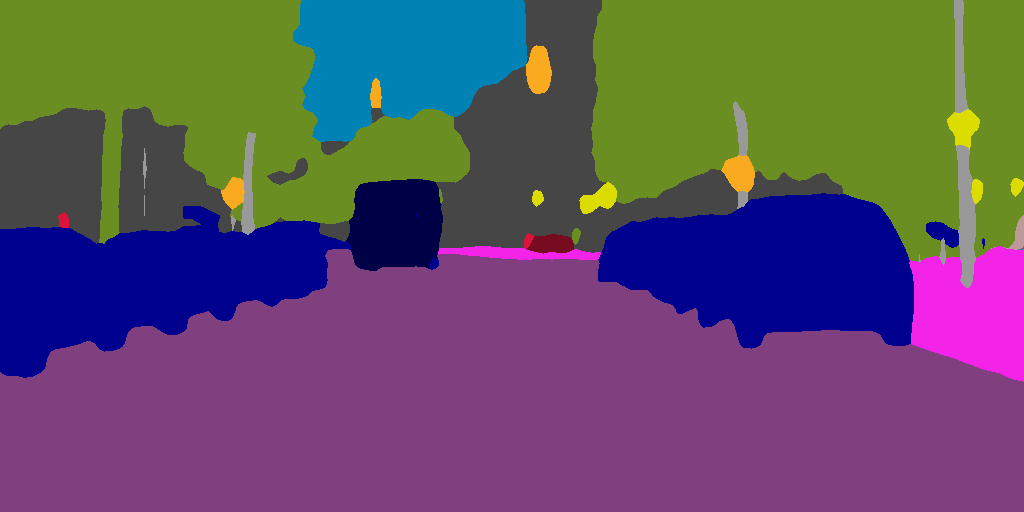}&
      \includegraphics[width=0.25\columnwidth]{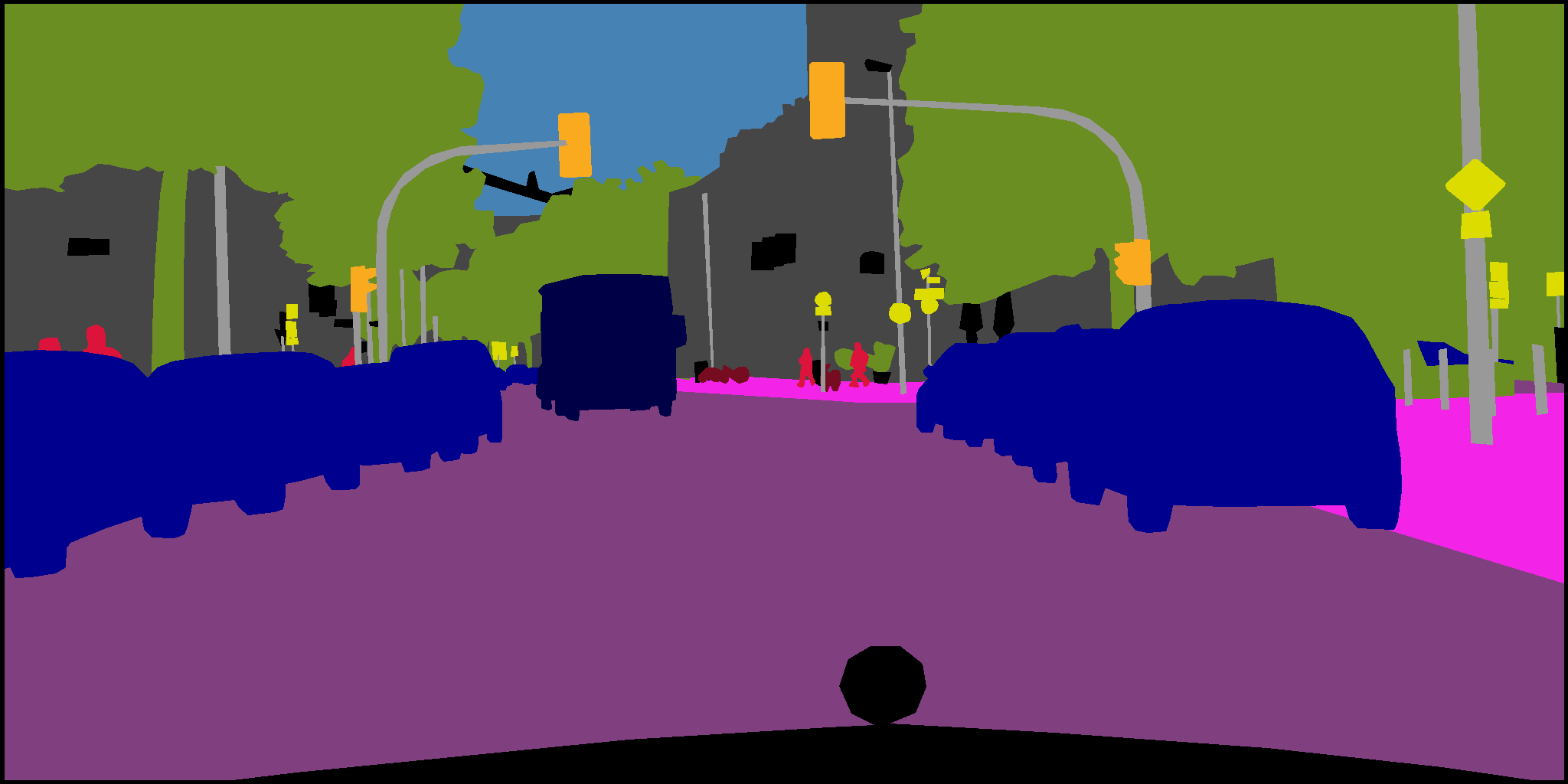}\\
  
      \includegraphics[width=0.25\columnwidth]{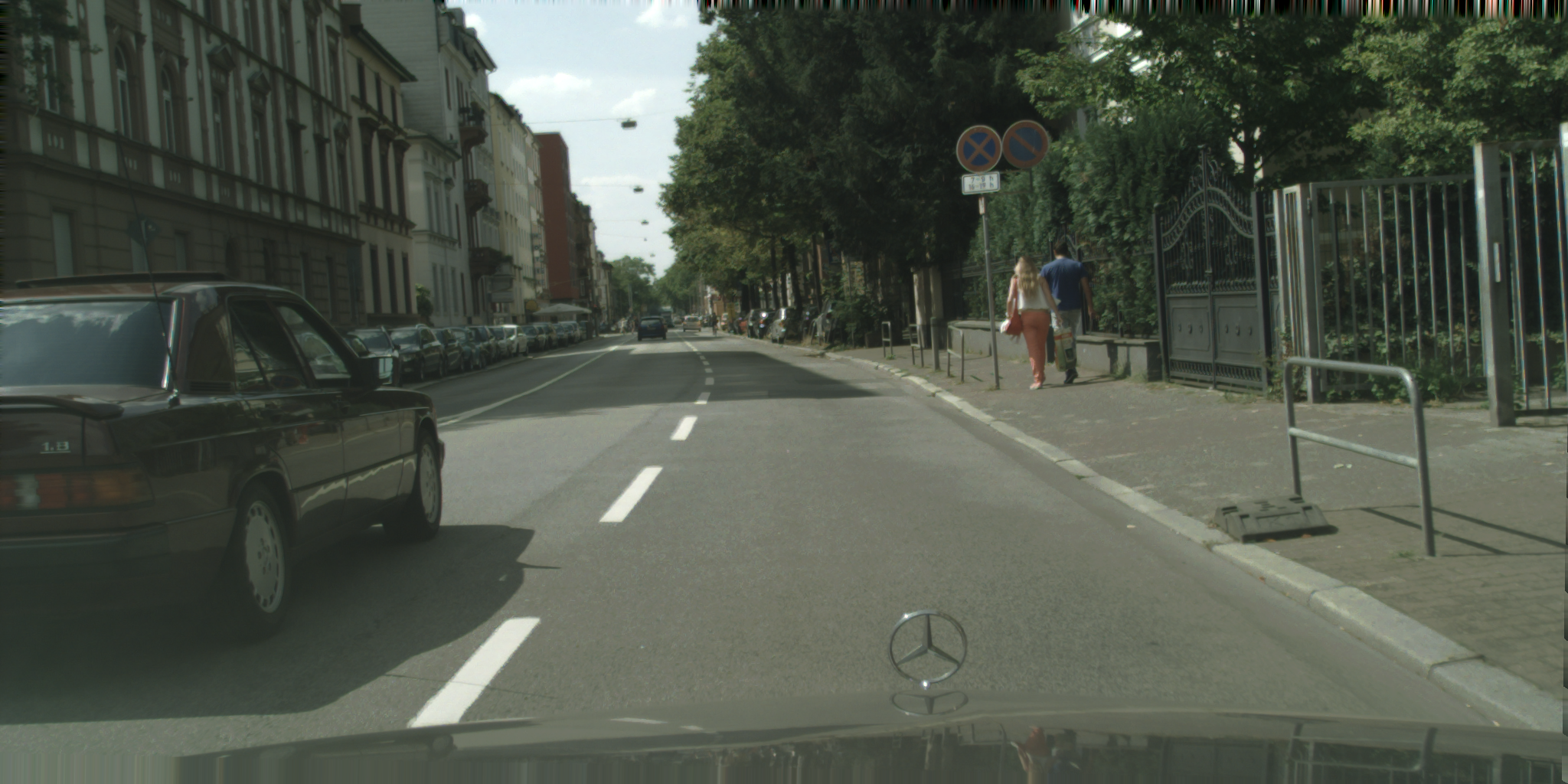}&
      \includegraphics[width=0.25\columnwidth]{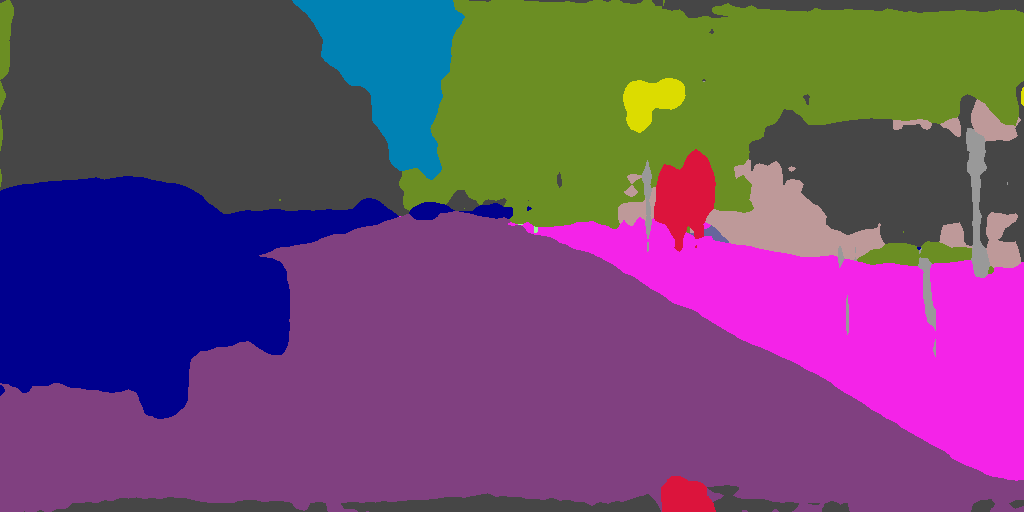}&
      \includegraphics[width=0.25\columnwidth]{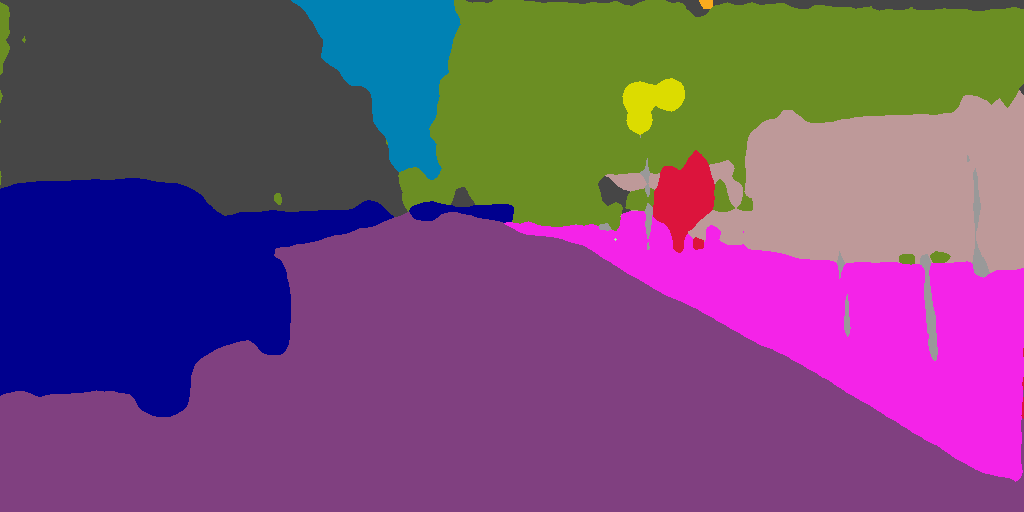}&
      \includegraphics[width=0.25\columnwidth]{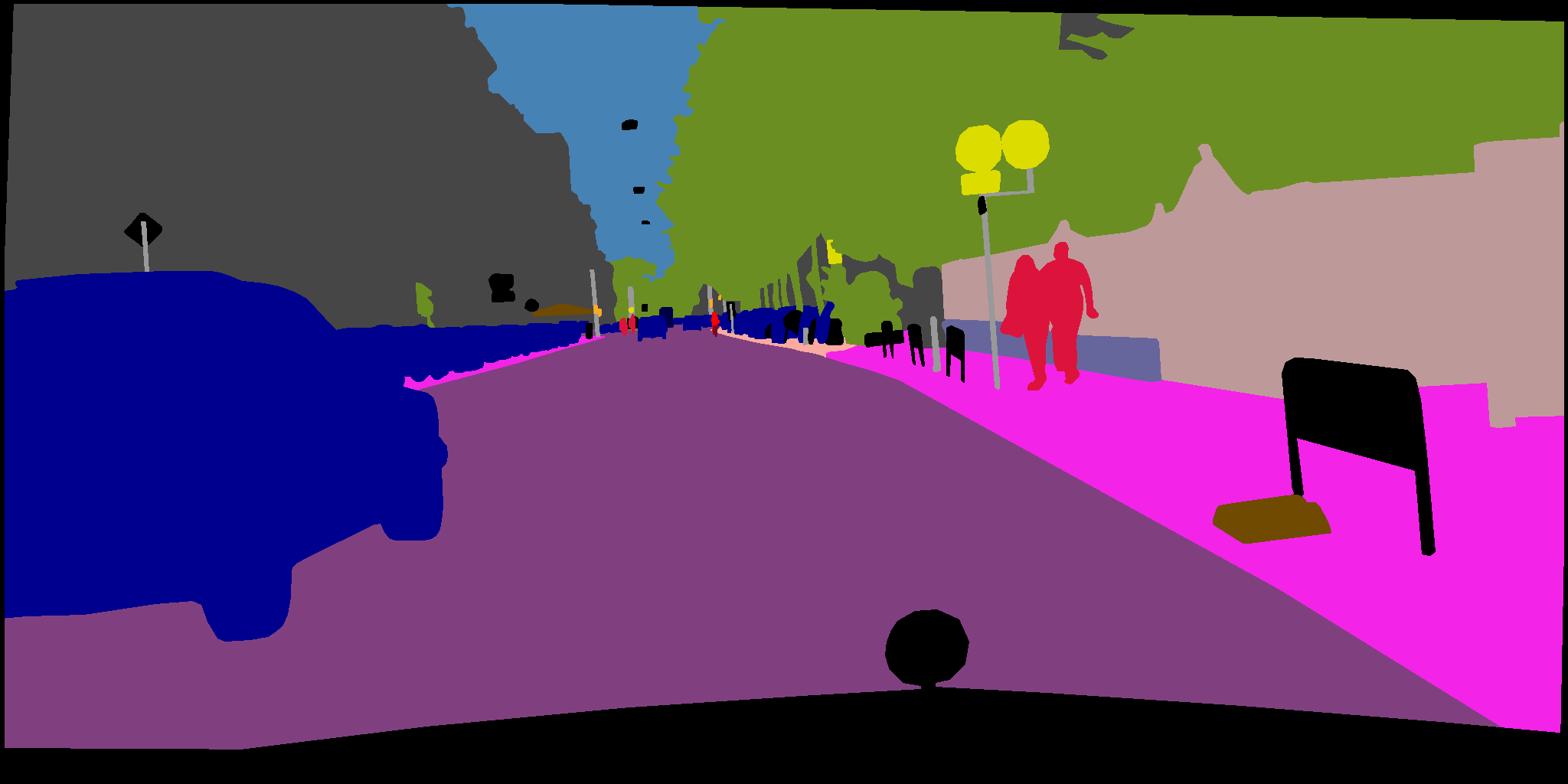}\\
      
      \includegraphics[width=0.25\columnwidth]{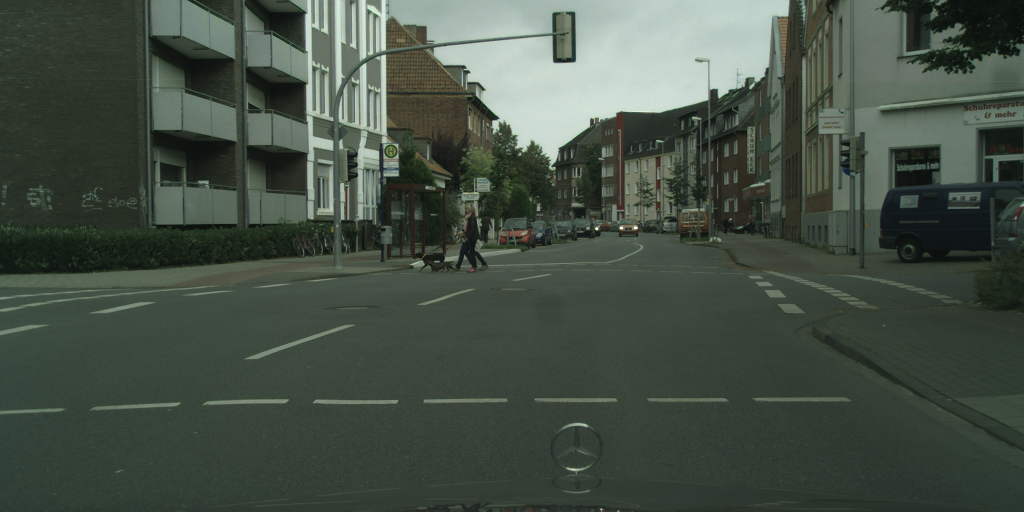}&
      \includegraphics[width=0.25\columnwidth]{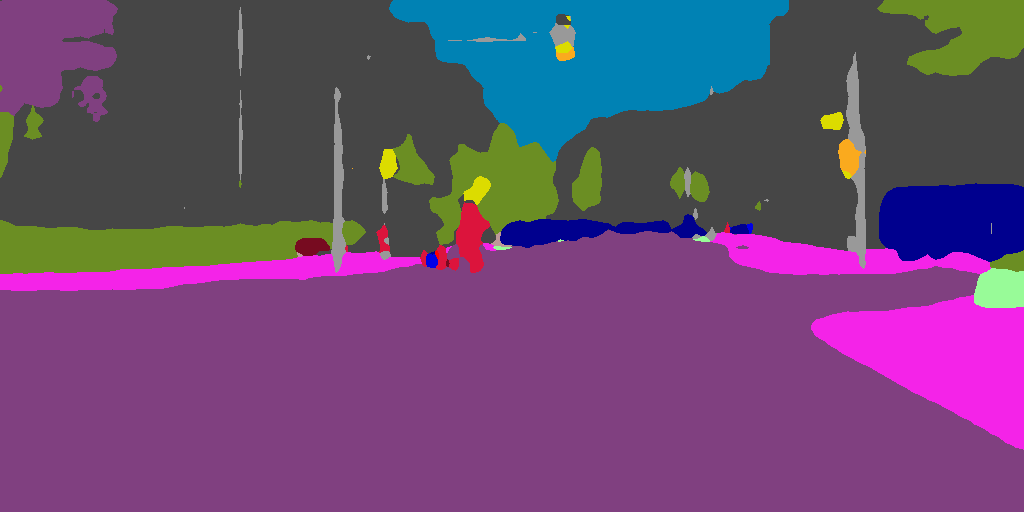}&
      \includegraphics[width=0.25\columnwidth]{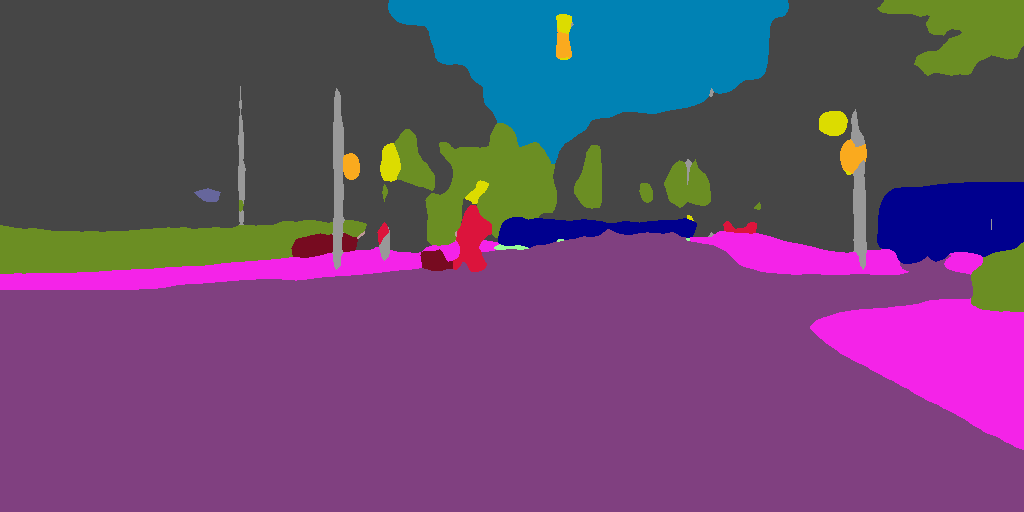}&
      \includegraphics[width=0.25\columnwidth]{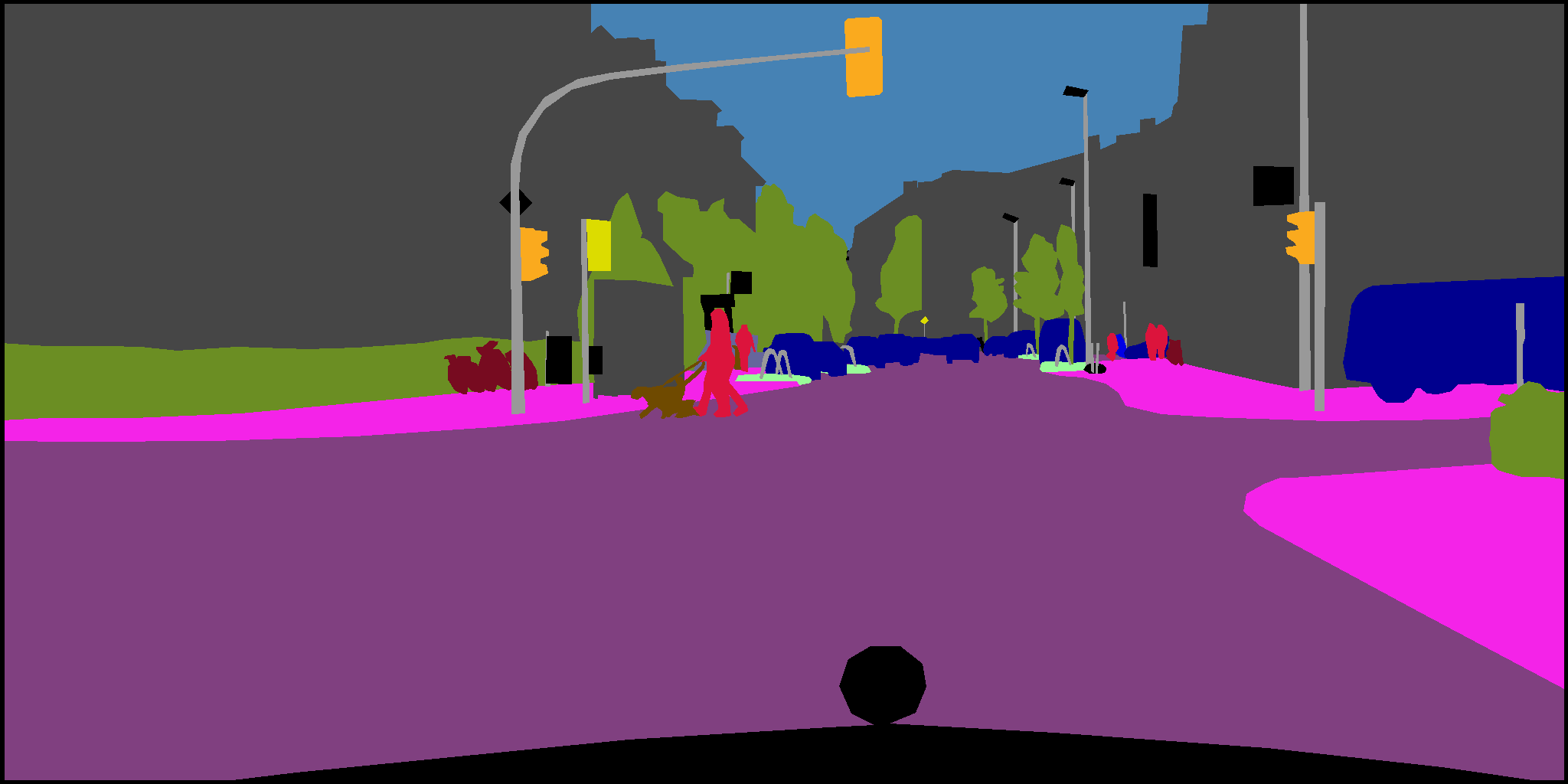}\\
      
      Image & CutMix~\cite{french2020semi} & Ours (RML) & Ground truth\\
  \end{tabular}
  }
  \caption{Qualitative comparison on Cityscapes dataset. Models are trained using 1/8 training images as labeled data.}
  \label{fig:qualitative1}
  \end{figure*}

\paragraph{The number of stages.}
We adopt a stage-wise training
strategy, i.e., once the mutual learning converges we recompute the pseudo labels, i.e., $p^0_k$ and then start the
next training stage. To show the effect of how the number of stages affect the performance,
we present the comparison in Table~\ref{table:stage}.
After training one stage, the mIoU is $62.85\%$ which is further improved to $63.65\%$ after training the second stage.
The third stage brings only $0.06\%$ gain which is marginal,
therefore our RML adopts two stages for the training process.

\begin{table}[!h]
  \footnotesize
  \centering 
  \caption{The performance on Cityscapes in different training stages (1/8 labeled images) .}
  \begin{tabular}{c|cccc}
  \toprule
  & Supervised baseline & Stage 1 & Stage 2 & Stage 3\\
  \midrule
  mIoU & 56.03 & 62.85 & 63.65 & 63.71\\
  \bottomrule
  \end{tabular}
  \label{table:stage}
\end{table}

\vspace{-0.2em}
\paragraph{Different thresholds.}
Prior pseudo labeling methods usually adopt a threshold to select some reliable predictions as pseudo labels. In our proposed robust mutual learning, we do not apply a threshold for convenient usage.
As a matter of fact, our approach can still make use of thresholding to remove some unreliable predictions. We experiment with different values of thresholds and present the comparison in Table~\ref{table:ablation}. 
Due to the powerful pseudo labels refinement, our RML is insensitive to the threshold chosen.
More importantly, the performance without thresholding is similar to that using thresholding,
suggesting that our approach is able to correctly rectify those unreliable pseudo labels.

\begin{table}[!h]
  \footnotesize
  \centering 
  \caption{Ablation on Cityscapes with different thresholds (1/8 labeled images, stage 1) .}
  \begin{tabular}{c|ccccccc}
  \toprule
  threshold & 0.0 & 0.2 & 0.4 & 0.6 & 0.8 & 0.9 & 0.95\\
  \midrule
  mIoU & 62.85 & 62.85 & 62.82 & 62.80 & 62.86 & 62.87 & 62.91 \\
  \bottomrule
  \end{tabular}
  \label{table:ablation}
\end{table}

\vspace{-0.2em}
\paragraph{Different confidence estimations.}
Our approach adopts the class-wise confidence that is estimated according to the feature distances to the class prototypes. 
One intuitive way of confidence estimation is to directly use the softmax output of the mean teacher, in which case, the function of updating the pseudo labels is changed from ${\bm{p}}^t_k = {\bm\omega}_k \cdot \bm{p}_k^{0}$ to ${\bm{p}}^t_k = {\tilde{h}_{{\bm{\theta}}}(\gamma(\bm{x}_u))} \cdot \bm{p}_k^{0}$. However, we conduct the experiments and observe that the performance drops from $62.85\%$ to $61.84\%$, demonstrating the effectiveness of prototypes. 

\vspace{-0.2em}
\paragraph{Class-wise confidence weights.} We assume equally weighted class prior when calculating the class-wise confidence ${\bm\omega}_k$ because we want to weight more for rare classes. Since there is a severe class imbalance in segmentation task, if we estimate the distribution according to class ratio, the model tends to ignore rare classes, such as traffic light. Hence, we empirically assume equally weighted mixture model. In this way, we implicitly rebalance the different classes, somewhat similar to the effect of focal loss. In our experiment, when we apply class weights in terms of class-wise pixel counts in the supervised set, the mIoU drops from 62.85 to 61.97.

{\small
\bibliographystyle{ieee_fullname}
\bibliography{egbib}
}